\RequirePackage{fix-cm}
\RequirePackage{mathtools}

\documentclass[twocolumn]{svjour3}         

\smartqed  

\usepackage{lmodern}
\usepackage{graphicx}
\usepackage[dvipsnames]{xcolor}
\usepackage{makecell}
\usepackage{cellspace}
\usepackage{setspace}
\usepackage[normalem]{ulem}
\usepackage{caption}
\usepackage{graphics}

\usepackage{amsmath}

\usepackage{amssymb}
\usepackage{latexsym}
\usepackage{multirow}
\usepackage{bbm}
\usepackage{enumitem}

\usepackage[utf8]{inputenc}
\usepackage[english]{babel}
\usepackage{hyperref}
\usepackage{comment}
\usepackage{subcaption}

\usepackage[numbers]{natbib}
\usepackage{algorithm}
\usepackage{algorithmic}
\usepackage[para,online,flushleft]{threeparttable}
\newcounter{savealgorithm}

\allowdisplaybreaks

\journalname{Health Care Management Science}

\begin{document}
\title{Surgical Scheduling via Optimization and Machine Learning with Long-Tailed Data}

\subtitle{\textit{Health Care Management Science}, in press}
\titlerunning{Surgical Scheduling via Optimization and Machine Learning with Long-Tailed Data}        %

\author{Yuan Shi\textsuperscript{*} \and Saied Mahdian\textsuperscript{*} \and Jose Blanchet \and Peter Glynn  \and \\ Andrew Y. Shin \and David Scheinker}
\authorrunning{Yuan Shi \and Saied Mahdian \and Jose Blanchet \and Peter Glynn \and Andrew Y. Shin \and David Scheinker}

\institute{\textsuperscript{*}Equal contribution \\ Yuan Shi is with Massachusetts Institute of Technology. \\ Saied Mahdian, Jose Blanchet and Peter Glynn are with Stanford University. \\ Andrew Y. Shin and David Scheinker are with Lucile Packard Children's Hospital and Stanford University.}
\date{November 20, 2022}
\maketitle
\begin{abstract} Using data from cardiovascular surgery patients with long and highly variable post-surgical lengths of stay (LOS), we develop a modeling framework to reduce recovery unit congestion. We estimate the LOS and its probability distribution using machine learning models, schedule procedures on a rolling basis using a variety of optimization models, and estimate performance with simulation. The machine learning models achieved only modest LOS prediction accuracy, despite access to a very rich set of patient characteristics. Compared to the current paper-based system used in the hospital, most optimization models failed to reduce congestion without increasing wait times for surgery. A conservative stochastic optimization with sufficient sampling to capture the long tail of the LOS distribution outperformed the current manual process and other stochastic and robust optimization approaches. These results highlight the perils of using oversimplified distributional models of LOS for scheduling procedures and the importance of using optimization methods well-suited to dealing with long-tailed behavior.


\smallskip

\noindent \textbf{Highlights}
\begin{itemize}[label = $\bullet$]
\item Cardiovascular post-surgical lengths of stay (LOS) are critical in optimizing recovery unit congestion, but extended LOS are very difficult to predict despite the use of a wide range of machine learning models and a rich set of patient characteristics.
\item Optimization models that rely on machine learning predictions of LOS without accounting for extended LOS did not improve scheduling performance (recovery unit congestion and wait times of patients) relative to current paper-based systems in use.
\item We show a data-driven conservative stochastic optimization approach that accounts for stochasticity in extended LOS can achieve scheduling performance improvements, outperforming other stochastic and robust optimization approaches.
\item We apply and evaluate our methodology in the context of a pediatric academic medical center using real medical and operational data.
\end{itemize}

\end{abstract}

\section{Introduction}

For hospital-based surgical care, the capacity of the intensive care unit (ICU) is often a crucial downstream bottleneck.  ICU bed shortages are associated with adverse patient outcomes, lost revenue from cancelled procedures, and a variety of detrimental spillover issues for numerous parts of the hospital. For many types of surgery, patients are sent to an intensive care unit (ICU) to recover until they meet the criteria to be transferred to a step-down unit. The capacity of the ICU, especially the number of beds available, is often a crucial bottleneck of surgical planning. When the ICU is at capacity, staff may transfer patients prematurely to the step-down unit or surgical procedures may be cancelled at the last-minute with adverse impact on patient and family experience, hospital reputation, finances and staff morale \cite{chan2012optimizing}. 

While most patients that require an ICU bed require it urgently, elective surgical procedures are scheduled as far as a year in advance. Our primary goal in this work was to develop a scheduling model that would reduce post-surgical bed congestion in practice in the presence of difficult-to-predict, long-tailed LOS data. Our secondary goals were to address the challenges associated with optimization in the presence of long-tailed empirical data. Numerous works have examined how to optimize surgical scheduling in order to reduce recovery-bed congestion. Despite the importance of post-surgical LOS, studies commonly discard empirical LOS data after they have been used to fit the parameters of a distribution and use only synthetic data to evaluate model performance. We illustrate limitations associated with using synthetic, rather than empirical, data.

\subsection{Setting}
\begin{figure}[!h]
    \centering
    \includegraphics[width = 1\linewidth]{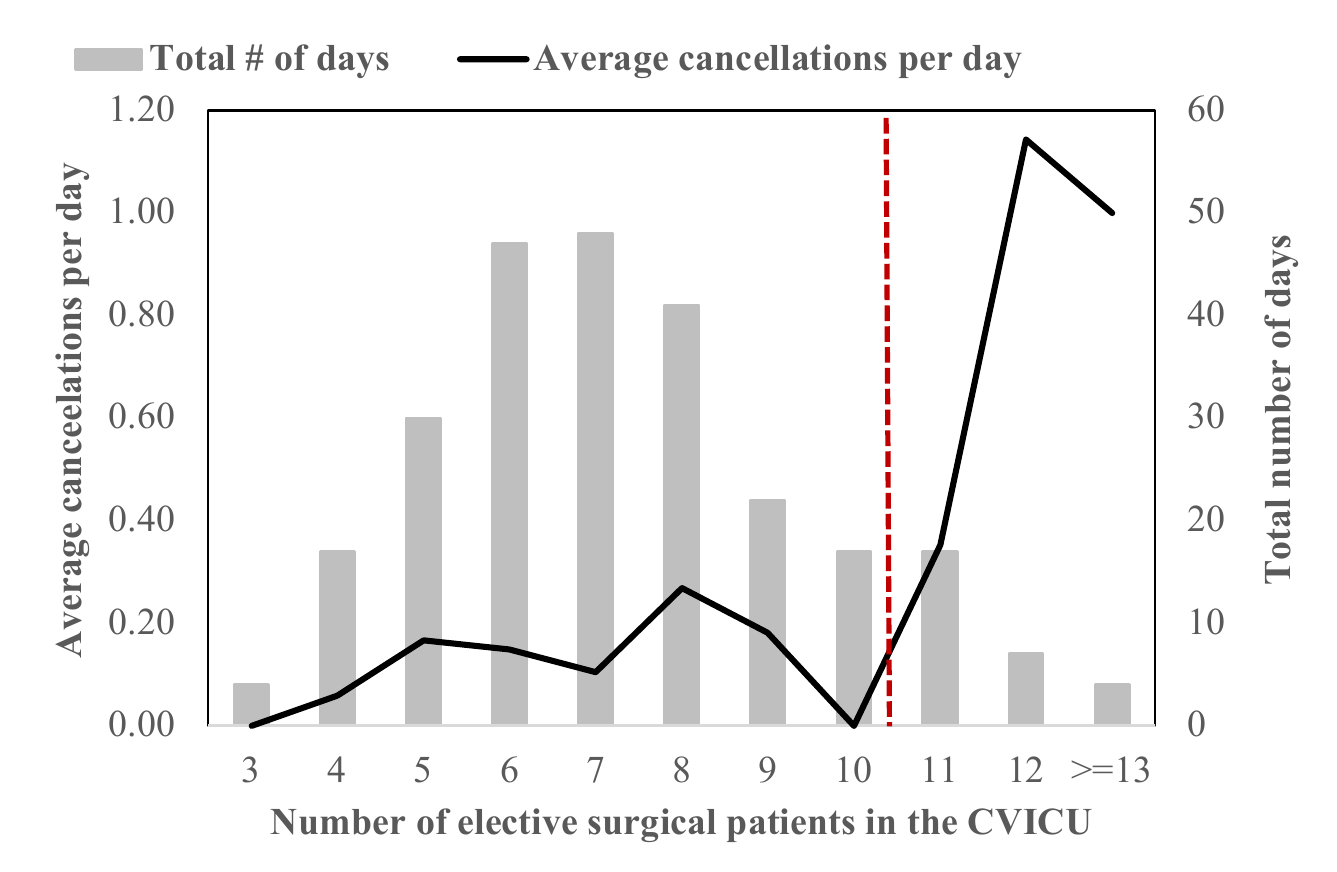}
    \caption{High ICU occupancy by elective patients is associated with high rate of surgical cancellations due to limited bed capacity}
    \label{fig1}
\end{figure}

This work was performed at the heart center of a high surgical volume pediatric academic medical center (PAMC) in the United States. The heart center uses three operating rooms, 26 cardiovascular ICU (CVICU) beds shared by elective and urgent surgical patients, and an acute care unit. In recent years, growing demand for elective surgical services and fixed CVICU capacity has resulted in numerous surgical cancellations. From September 2019 to May 2020, 84\% of cardiovascular surgery cancellations happened on the 40\% of days with 23 or more patients in total in the CVICU. In particular, 35\% of cancellations happen on the 11\% of days with more than 10 elective surgical patients in the CVICU (See Figure \ref{fig1}). This suggests that a significant fraction of cancellations may be prevented if optimized scheduling smooths elective surgical patient CVICU occupancy.

\subsection{Overview}
Numerous theoretical works using optimization to schedule elective surgical cases have demonstrated substantial reductions to ICU congestion in numerical experiments \cite{min_yih,robust_opt,zhang2019two,dro}. However, relatively little work has been implemented or has demonstrated measurable improvements in practice. Most existing positive results rely on the use of synthetic data, e.g., a log-normal distribution fit to approximate empirical data from the institution studied, to represent procedure duration and post-operative length of stay. Such common practice, combined with the lack of real-life implementations of these algorithms, leaves many real world challenges of surgical scheduling unaddressed and unidentified. Meanwhile, many works have examined empirical data in order to predict the length of stay (LOS) of patients in the ICU (see, for example, \cite{weissman2018inclusion,ReviewEttema10,Nassar16physician}), but little has been done in utilizing such LOS predictions for surgical scheduling.

We examine reducing ICU congestion by optimizing the scheduling of complex pediatric cardiovascular surgical procedures, using real data from the hospital for patient flow and post-operation ICU LOS. Our approach combines machine learning for LOS forecasting, optimization for surgical scheduling, and simulation for performance evaluation. 

To motivate the problem, we first establish the theoretical upper bound of system performance by running a deterministic optimization formulation with known \textit{a priori} information on post-operation LOS. The results demonstrate significant reductions in CVICU congestion and patient wait times, compared to the institution's current manual and primarily paper-based process. We then develop predictive models using machine learning for LOS based on patient data available at the time the procedure is scheduled. When perfect information of LOS is replaced with point-predictions using machine learning, however, deterministic optimization is found to be not sufficient to reduce congestion without increasing wait times. This is despite access to a very rich set of patient characteristics for LOS prediction. The key bottleneck is identified to be the lack of accurate predictions of LOS, especially for a small group of patients with very long LOS. This has a disproportionately large impact on optimization performance due to the long tail of the distribution of LOS. We observe that patients with similar characteristics may vary drastically in realized LOS, and patient populations display temporal non-stationarity, both suggesting inherent unpredictability of LOS at the time of scheduling. This inspired the development of stochastic and robust optimization algorithms that incorporate LOS probability distributions instead of point estimates to address this bottleneck.


We develop a novel data-driven modeling framework for scheduling cardiovascular surgery under uncertain LOS. The framework combines machine-learning predicted LOS probability distributions, rolling information update and optimization methods. Under this framework, three scheduling algorithms are formulated: two based on stochastic optimization, namely Standard-RSO and Conservative-RSO, and one robust optimization formulation, RRO.  
The optimal schedules given by each algorithm are evaluated through simulation using historical patient arrivals and LOS. 

Machine learning models are used to obtain both point-estimates of LOS and its probability distribution characterized by predictive errors. Standard-RSO, which considers both under- and over-estimation errors, fails to reduce ICU congestion versus the status quo. In contrast, both Conservative-RSO and RRO, with carefully tuned parameters, managed to reduce ICU congestion without excessively increasing average patient wait times. We attribute the promising performances of the latter two algorithms to their targeted design focusing on addressing common under-estimation for prolonged LOS by machine learning predictions. Moreover, Conservative-RSO achieves better average and worst-case performance compared to RRO in reducing ICU congestion, revealing that algorithmic complexity does not always equal better performance in practical settings.


The main contributions of our work are two fold. First, the negative results associated with deterministic optimization as well as Standard-RSO offer important lessons that are generally applicable to others seeking to optimize surgical scheduling for complex patients. We highlight the importance of evaluating model performance using empirical data or simulation that fully captures the long-tailed nature of patient LOS. In addition, as hospital operations and medical practices are constantly changing, non-stationarity in LOS should be taken into account when developing simulating and evaluating data-driven models. Our negative results, revealed through simulation with empirical data, provide clear evidence against extrapolating good performance on synthetic data using standard distributional assumptions (e.g. lognormal distribution) into real world scenarios. These findings may also help explain the dearth of studies reporting successful implementation and measured improvements of similar approaches. 

Our second contribution is to propose a promising algorithm, Conservative-RSO, that is specifically designed to address the challenge of long-tailed ICU LOS in the surgical scheduling problem. Compared to Standard-RSO, Conservative-RSO demonstrates that good performance can be achieved by incorporating careful design choices without increasing computational complexity; the comparison with RRO further suggests that computationally complicated models do not always translate into better performance in practice. We believe the promising performance of Conservative-RSO helps identify a direction of future research on data-driven optimization methods to address one of the most challenging obstacles in hospital operations.

The remainder of the paper is structured as follows. Section \ref{section:related_work} provides a review of related literature on surgical scheduling optimization and LOS predictions. Section \ref{sec:determ} presents our modeling framework under deterministic LOS estimations, introducing formulations for both offline optimization and rolling optimization. These deterministic models are used to establish performance upper-bounds given accurate predictions of patient LOS. Section \ref{sec:los_predict} presents the development of machine learning models for LOS predictions and discusses the poor performance and challenges of using deterministic optimization with machine learning prediction. Section \ref{sec:sp} presents our modeling framework using predicted LOS distributions, and introduces the three algorithms, Standard-RSO, Conservative-RSO and RRO. The performances of these algorithms under numerical experiments are presented in Section \ref{sec:results}. Section \ref{Insights} discusses insights from the design and performance of algorithms, as well as their limitations and potential extensions. Finally, Section \ref{Conclusion} provides concluding remarks on implications of the work, best practices in applying schedule optimization in practice, and directions for future research.

Gurobi 9.0.3 \cite{gurobi} and Python were used to implement and solve all mixed integer optimization algorithms developed in the paper. All optimization problems were solved using the compute facility of Stanford Research Computing Center and Stanford University. On this shared facility, each optimization problem was solved with two CPUs and with either 8GB or 16GB of memory for each CPU.

\section{Related Work}\label{section:related_work}

\subsection{Surgery Scheduling}
Extensive research has been carried out on scheduling of patients to improve operating room (OR) performance. Most of the literature on elective surgery scheduling focuses on OR room capacities without considering the capacity of the subsequent recovery units. The variability of surgical procedure durations is usually considered as the primary source of stochasticity and as the primary challenge to surgical scheduling. We refer to \cite{zhu2019operating,rahimi2021comprehensive} for a comprehensive review and focus on research that studies the surgical scheduling problem under the constraint of limited downstream capacity.

Deterministic formulations of surgical schedule optimization under limited downstream capacity have been examined in \cite{hsu03,Guinet03,Pham08,fairley_scheinker}. \cite{hsu03} considered optimizing surgical scheduling with limited capacity at the Post Anesthesia Care Unit (PACU) to minimize the number of nurses needed for the PACU. The case duration and the recovery times in the PACU are treated as deterministic. 
Similarly, \cite{Pham08} considered the surgical scheduling problem with deterministic recovery times in the PACU and the ICU for all patients under the assumption of perfect information.
\cite{Guinet03} also investigates surgery scheduling with downstream PACU capacity constraints. It proposes a two step optimization procedure where patients are first assigned to ORs and dates. In the second step  surgery times are assigned to patients.  
More recently, \cite{fairley_scheinker} developed a combined machine learning and optimization approach to reduce congestion in the PACU, where the recovery time in the PACU is estimated using machine learning predictions. 
\cite{fairley_scheinker} reports good results based on simulations using empirical data, but does not report the results of implementation. Most of the existing deterministic formulations focus on the PACU capacity as the key downstream bottleneck with limited discussions on the ICU capacity. Compared to the recovery time at the PACU (which is typically in the order of hours), ICU LOS may be multiple days or months and involves greater variance than single-day PACU stays. 
The substantial difference in variability between CVICU long-tailed LOS and PACU recovery time are why the approach in \cite{fairley_scheinker} is not applicable in the present setting. 

Given the multi-period nature and significant uncertainties involved in ICU capacity planning, a number of mathematical formulations have been proposed in recent years. \cite{min_yih} provided a well-known benchmark instance of surgery scheduling with ICU capacity constraints using a stochastic mixed integer programming model. The work uses sample average approximation and assumes the LOS in the ICU to be random with known distributions (arbitrary with finite support).
\cite{zhang2019two} proposes a two level time horizon for surgery scheduling. In the first level, patients are selected from a waitlist to be scheduled with a timeframe (e.g. a week) using approximate dynamic programming. In the second level, selected patients are scheduled using a sample average approximation method similar to \cite{min_yih}. 
Other stochastic programming approaches built on the work in \cite{min_yih}; we refer to \cite{dro} for a review of these formulations.

Besides stochastic optimization models, \cite{robust_opt,dro} develop optimization methods for robust performance of surgery scheduling under worst-case realizations of LOS or LOS distributions. Most relevant to our work is \cite{robust_opt}. The authors formulated a two-stage robust optimization approach to reduce congestion in downstream capacities, and developed solution techniques which we adapt and apply to our setting.

All of the above-mentioned papers on stochastic or robust optimization evaluated model performance using synthetic data based on strong distributional assumptions, and none has been implemented in practice. In contrast, we use real LOS data in evaluating our model performance.

In addition, \cite{fugener14} considered a surgery scheduling problem focusing on minimizing downstream costs including overcapacity costs at the ICU. Models for this problem proposed in \cite{fugener14} and others such as \cite{belien2009decision} are concerned with the \textit{tactical} problem of allocating OR block of times to surgical specialties to optimize patient flow into the downstream units. In contrast, our work and others' mentioned above focus on the \textit{operational} problem of assigning individual patients to surgical time blocks.

To our knowledge, we are the first to study a combined machine learning and optimization approach for schedule optimization to reduce congestion at the CVICU: an environment that includes non-stationary, long-tail LOS behavior. We are also the first to propose a stochastic formulation for surgical scheduling that is specifically designed to address significant variability in the tail of the LOS distribution. We use real world post-operation LOS data for LOS prediction, model parameter tuning and performance evaluation, reflecting the difficulties of working in a setting with non-stationary operations and patient volumes and long-tailed LOS distributions.

\subsection{Post-surgery LOS prediction}\label{sec:review-pred}
In developing our machine learning models for LOS prediction, we refer to a separate line of literature focusing on predictions of post-surgery (in particular cardiac surgeries) LOS in the intensive care units. 

Most work suggests that post-surgery LOS prediction at admission time - either using predictive modeling or by expert opinion - is  challenging especially in the case of prolonged ICU LOS. 

One common way to predict prolonged ICU LOS is through binary classification, such as in \cite{weissman2018inclusion,ReviewEttema10}. However, binary classification does not provide the level of granularity required for optimizing surgical scheduling in our context. 

For regression-based models, \cite{whellan2011predictors} builds and evaluates multivariate regression models using data from 246 hospitals for heart failure patients. The model achieves a modest $R^2$ value of 4.8\%, where only 1.2\% of variation is explained by patient characteristics. Similarly, \cite{chen1994variation} shows through univariate regression that only 12\% of the variation could be explained by patient characteristics and general hospital characteristics in aggregate for patients with a primary diagnosis of acute myocardial infarction. More complicated LOS regression models using machine-learning for cardiac surgery patients are artificial neural networks developed in \cite{nnLOS,lafaro2015neural} and adaptive neuro-fuzzy systems explored in \cite{maharlou2018predicting}. Although machine-learning based models generally result in a higher $R^2$ value, predictive accuracy for patients with prolonged LOS remains low. For instance, the neural network in \cite{nnLOS} achieves an overall accuracy of over 60\% but is unable to predict any LOS above 15 days, and the model in \cite{lafaro2015neural} with $R^2=0.41$ underestimates the LOS for almost all patients with actual LOS of longer than 100 hours. 

More generally, \cite{Gusma04physician,Nassar16physician} demonstrate that prolonged LOS predictions are challenging even for experienced physicians. In addition, \cite{kramer2010predictive} suggests that information gathered at admission did not have a significant impact on the identification of patients with prolonged ICU LOS. The variables that had the greatest impact on prolonged ICU LOS were those measured on day 5 of ICU stays. Although LOS prediction using patient features collected during and after surgery tends to achieve a higher level of accuracy \cite{kramer2010predictive,tabbutt2019novel,yang2010predicting}, most of these data are not available when a procedures is being scheduled weeks to months in advance.

In our attempt to develop a predictive model for post-operation LOS, we identified the same difficulty as observed by the others in dealing with the long-tailed, non-stationary distribution of LOS for complex cardiovascular surgeries. This challenge motivated our design of a stochastic optimization formulation that specifically addresses the unpredictability and the consistent underestimation of prolonged LOS by predictive models. We discuss the formulation in detail in Section \ref{sec:sp}.

\section{Deterministic Formulations for Schedule Optimization}\label{sec:determ}

We first study the problem of surgical scheduling where the LOS in the ICU are treated as deterministic. We develop two optimization models - offline and rolling deterministic optimization - given operational constraints at the PAMC and some \textit{a priori} information on patient LOS. 

\subsection{Offline Deterministic Optimization}\label{section:offline}

We formulate the offline surgery scheduling problem as a mixed integer program (MIP) with the objective of smoothing out CVICU elective census overtime to reduce cancellations without creating excessive delays. 

Define set $D = \{1,2,\ldots,N_d\}$ as the set of available dates in the time period of interest, and define set $P = \{1,2,\ldots,N_p\}$ as the set of all elective surgical patients that are to be scheduled within the time period. We define binary decision variables, $$x=\{x_{d,p}:d\in D,p\in P\},$$ where $x_{d,p}=1$ if patient $p$ is scheduled for surgery on day $d$ and otherwise $x_{d,p}=0$.

For the offline problem, it is assumed that both patient arrivals, $P$, and the LOS of each patient, $l_p$, are known in advance. Given $x$ and $\{l_p:p\in P\}$, the ICU overflow variable, $u_d$ for all $d\in D$, counts the number of elective patients on day $d$ that exceed ICU capacity, $c$.

The offline deterministic optimization problem is formulated as below. 
\begin{subequations}
\begin{equation}
    \min_{x}\quad \sum_{p\in P}
    \sum_{d=1}^{N_d}(d - d^{min}_p)^+ x_{d,p} + \beta \sum_{d=1}^{N_d} f(u_d) 
    \label{opt1:objective}
\end{equation}
\begin{equation}
    \sum_{d \in D} x_{d,p} = 1\quad \forall p\in P \label{con:match}
\end{equation}
\begin{equation}
\begin{aligned}
y_{d,p} = \sum_{d'=\max(d-l_p+1,1)}^d x_{d',p}, \quad \forall p\in P, d\in D
\end{aligned}
\label{con:y}
\end{equation}
\begin{equation}
    \sum_{p\in P} y_{d,p} \leq c + u_d \quad \forall d\in D \label{con:overflow}
\end{equation}
\begin{equation}
 x\in Q^{op}  \label{con:operations}
\end{equation}
\begin{equation}
   y_{d,p},x_{d,p}\in\{0,1\},\quad u_d\geq 0\label{con:vars}
\end{equation}
\end{subequations}
We use the notation $(\cdot)^+ = \max(\cdot, 0)$.

In the offline problem, the first term of the objective function~\eqref{opt1:objective} represents the total wait time for all patients. Here, $d_p^{min}$ denotes the earliest date by which patient $p$ will be ready for his surgery. The second term, $\beta \sum_{d\in D} f(u_d)$, is the total cost associated with ICU overflow, weighted by constant $\beta$. We set the function $f(\cdot)$ to be a convex, piece-wise linear function of $u_d$ such that it approximates the quadratic cost function, $u_d^2$. The purpose of the convex cost function is to impose higher penalty for greater overflow in order to smooth out and minimize large peaks in ICU occupancy. This also reflects the highly non-linear increase in the possibility of cancellation as elective ICU occupancy increases, as demonstrated in Figure~\ref{fig1}. We provide the exact MIP formulation of $f(\cdot)$ in Appendix~\ref{app:math_formulation}.

Constraint~\eqref{con:match} enforces the assignment of every patient to exactly one date of surgery. The second constraint~\eqref{con:y} calculates if a patient needs an ICU bed on a given day based on her assignment and LOS. The third constraint~\eqref{con:overflow} calculates the ICU overflow, $u_d$, on each day given ICU bed capacity, $c$. 

For the last constraint~\eqref{con:operations}, we use $Q^{op}$ to represent any remaining institution-specific operational constraints. Scheduling for complex procedures is commonly constrained by single surgeon-patient matches, surgeon and OR room availability, which may differ depending on the context. For our formulation, $Q^{op}$ is tailored to the context of the PAMC as explained below.

\textit{Patient availability.} 
When scheduling patients for surgery, there is often a clinically determined upper bound on patient wait time and also a corresponding lower bound that accounts for surgical delays due to patient availability, travel arrangement and necessary insurance or health checks. We incorporate this constraint by imposing windows of availability for individual patients as hard constraints in our formulation, similar to \cite{Guinet03}. In practice, the window of availability of patients can be estimated case by case upon arrival and hence used as inputs for optimization.
    
For the purpose of simulation, since a patient's actual window of availability at the time of arrival is not recorded in historical census and surgery data from the PAMC, we use the following heuristics to obtain an estimation. The earliest available date of patient $p$ for surgery, denoted as $d_p^{min}\in D$, is set to be date that is half of the lead time prior to her originally scheduled surgery date. The latest available date, denoted as $d_p^{max}\in D$, is set to be 90 days after the actual surgery date. This heuristic ensures that the set of feasible scheduling solutions is not too restricted, and any resultant increase in wait time will be penalized by the first term in the objective function.

\textit{Surgeon and OR Room availability.} Each procedure requires one OR room and takes a pre-specified amount of time for the patient's assigned surgeon to complete. While many past work focused their attention on the stochastic nature of surgical durations (see, for example, \cite{robust_opt}), OR room availability is not a bottleneck resource in our setting. We therefore treat surgical durations as deterministic and uses the actual length of procedure in optimization\footnote{Alternatively, we also estimated each patients' surgical duration using the average surgical duration for each procedure type in the empirical data. This approach produces similar results to that using the actual surgical durations.}. 

Specific to the PAMC, a maximum of 45 hours of surgery in OR is available on every weekday, where each surgeon performs operations for no longer than 15 hours every day\footnote{Surgeries that took longer than 15 hours in reality were treated as 15 hours for feasibility.}. During implementation, we also incorporate surgeon-specific day-offs in the formulation. For example, one of the surgeons does not perform surgeries on Mondays.

\textit{Complex surgeries.} On top of the regular capacity constraints above, special arrangements are often required for complex, full-day procedures. At the PAMC, Pulmonary Artery Reconstruction (PAR) surgery is a type of highly complex procedure that requires special treatment in the algorithm. The surgeons only perform PAR surgeries on certain weekdays (`PAR days') and not the others. If a PAR surgery is scheduled for a surgeon on any day, no other surgery can be performed by the same surgeon on the same day. In addition, the surgeons need to take at least a one-day break in between PAR surgeries, i.e., PAR surgeries cannot be scheduled on two consecutive days for the same surgeon.

Full mathematical formulations of $Q^{op}$ for our context and implementation details are provided in Appendix~\ref{app:math_formulation}. 

Here, the weight for ICU overflow, $\beta$, and ICU bed capacity, $c$ are parameters to be tuned. For the value of $\beta$, we select a value among 
\begin{equation*}
\beta\in\{5,10,25,50,100,200\}
\end{equation*}
so that the model achieves the the most reduction on ICU overflow without excessively lengthening the average and median patient wait times compared to the original surgical schedule. 
Meanwhile, we determine the value of $c$ based on the minimum achievable ICU occupancy upper bound given true historical on LOS, i.e.,
\begin{equation}
    c = \min_{x} \max_{d\in D}\quad \sum_{p\in P} y_{d,p}
    \label{c_objective}
\end{equation}
\begin{equation*}
    \text{s.t.}\quad \text{Constraints~\eqref{con:match}, \eqref{con:y}, \eqref{con:operations}, and \eqref{con:vars}}
\end{equation*}
Solving~\eqref{c_objective} yields $c=8$ at the PAMC, which is used for all numerical experiments for the rest of the paper.

\subsection{Rolling Deterministic Optimization (RDO)}\label{sec:batch}
In practice, the arrival and LOS of incoming patients are unknown and performing offline optimization over a one-year horizon is not practical. We thus develop a rolling-horizon alternative to the offline formulation, where patients are scheduled in batches with estimated LOS at the time of scheduling. Meanwhile, we dynamically observe the realization of LOS for patients who have undergone surgery in the previous period and update LOS estimates for these patients.

We define a sequence of scheduling days, $s_b\in D$, for $b\in\{1,2,\ldots,B\}$ in chronological order. On each scheduling day $s_b$, scheduling is performed for all patients who arrived between $s_{b-1}$ and $s_{b}-1$. We use $P_b$ to denote the batch of patients who are scheduled on day $s_b$. These patients may be scheduled for surgery any time during the scheduling horizon, $s_b, s_b + 1, \ldots, s^{max}_b$, where $s^{max}_b\in D$ is the latest date that any patient in $P_b$ can be scheduled for surgery, i.e., $s^{max}_b = \max_{p\in P_b} d^{max}_p$. To simplify implementation, we set $s^{max}_b = N_d$ for all $b$, where $N_d$ is sufficiently large to cover the scheduling horizon for all batches.\footnote{Note that we still enforce the constraint that each patient is scheduled no later than $d^{max}_p$ through $Q^{op}$.}

Note that scheduling decisions for patients in $P_b$ are also influenced by past patients, $P^{past}_b = \bigcup_{k=1}^{b-1}P_k$ who are still in the system. This includes all patients who arrived before day $s_{b-1}$ who are currently staying in the ICU on day $s_b$, as well as those who have their surgery scheduled for days within the current scheduling horizon. Surgical and ICU capacities are adjusted accordingly to account for these past scheduling decisions. 

To introduce the mathematical formulation, we use the same set of notations as the offline problem. Note that the binary decision variables $x_{d,p}$, $y_{d,p}$ are now defined for all $d\in D$ and $p\in P_b \cup P^{past}_b$. Since scheduling decisions for patients in $P^{past}_b$ are already made, we let constants $\tilde{x} = \{\tilde{x}_{d,p}: d\in D, p \in P^{past}_b\}$ denote previous scheduling decisions. This constraint implicitly updates surgical and ICU capacities available for incoming patients by accounting for patients who have been previously scheduled. The deterministic batch optimization problem (BOP) on scheduling day $s_b$ is formulated as below. 
\begin{subequations}\label{eqns:deterministic BOP}
\begin{equation}
    \min_{x}\quad \sum_{p\in P_b}
    \sum_{d=s_b}^{N_d}(d - d^{min}_p)^+ x_{d,p} + \beta \sum_{d=s_b}^{N_d} f(u_d) \label{opt2:objective}
\end{equation}
\begin{equation}
x_{d,p} = \tilde{x}_{d,p}\quad \forall p\in P^{past}_b, d\in D
\label{con:past}
\end{equation}
\begin{equation}
    \sum_{d=s_b}^{N_d} x_{d,p} = 1,\quad \sum_{d=1}^{s_b -1} x_{d,p}=0 \quad \forall p\in P_b \label{con:match2}
\end{equation}
\begin{equation}
\begin{aligned}
y_{d,p} = \sum_{d'=\max(d-l_p+1,1)}^d x_{d',p}  \quad \forall p\in P_b\cup P_b^{past}, d\in D
    \label{con:y2}
\end{aligned}
\end{equation}
\begin{equation}
    \sum_{p\in P_b \cup P^{past}_b} y_{d,p} \leq c + u_d \quad \forall d\in D \label{con:overflow2}
\end{equation}
\begin{equation}
x\in Q^{op}  \label{con:operations2}
\end{equation}
\begin{equation}
   y_{d,p},x_{d,p}\in\{0,1\},\quad u_d\geq 0\label{con:vars2}
\end{equation}
\end{subequations}

The objective function~\eqref{opt2:objective} mirrors that of the offline formulation in Equation~\eqref{opt1:objective}. Constraint~\eqref{con:past} requires that patients scheduled previously are not re-scheduled for a different date. Constraint~\eqref{con:match2} requires that every patient in $P_b$ is scheduled for one surgery day no earlier than $s_b$. Constraints ~\eqref{con:y2}-\eqref{con:vars2} mirror constraints \eqref{con:y}-\eqref{con:vars} in the offline problem. Note that constraints \eqref{con:overflow2} and \eqref{con:operations2} are influenced by both patients in $P_b$ and those in $P^{past}_b$. This is because the latter may also undergo surgery and/or need ICU beds after the scheduling date $s_b$, thus competing for ICU resources and other resources in $Q^{op}$ (e.g. surgeon and OR room availability) with patients in $P_b$. Full mathematical formulation with our context-specific $Q^{op}$ and implementation details are provided in Appendix \ref{app:batch_deterministic_formulation}.

Surgical schedule optimization can thus be operationalized in practice by solving deterministic BOP sequentially for $b=1,2,\ldots, B$, with an appropriately chosen sequence of scheduling days, $\{s_b\}$ and estimates of patient LOS, $\{l_p:p\in P_b\cup P^{past}_b\}$. For patients in $P_b$, estimations of $l_p$ can be made based on patient features at the time of arrival (see Section~\ref{sec:los_predict}). For patients in $P^{past}_b$ who have received their surgeries, $l_p$ estimates can be progressively updated as uncertainties in LOS realize. We formalize this information update procedure below.

\paragraph{Information Update Procedure for deterministic BOP.} At the start of each batch $b$, the value of $l_p$ for all $p \in P^{past}_b$ is updated as follows.
    \begin{enumerate}[label=(\alph*)]
        \item  If the patient has undergone surgery and has been discharged from ICU by scheduling day $s_b$, her realized LOS is observed and we update $l_p$ to be the true LOS.
        \item If the patient is in the ICU on day $s_b$ having stayed for $m$ days, then $l_p$ is updated based on partially realized LOS and any additional post-operations information. 

    \item If the procedure of the patient is scheduled after $s_b$, there is no change to $l_p$.
    \end{enumerate}
For simulation purpose only, we use a simplified heuristic for patients under (b). We assume the true LOS can be evaluated with certainty for those soon to be discharged (within 5 days), and thus update $l_p$ to be the true LOS for those patients. For the remaining patients, we set $l_p \leftarrow \max\{m+10,l_p\}$. In practice, $l_p$ can be re-evaluated on a case-by-case basis for each patient based on any new information available. 

On every scheduling day, patient arrivals $P_b$ and any realization of additional LOS information for $P^{past}_b$ are observed. An optimal solution of deterministic BOP is then obtained and implemented for all patients in $P_b$. The existing schedule and required surgical resources are updated accordingly. We describe this rolling optimization process in Algorithm~\ref{RDO}.

\begin{algorithm}
\caption{\small Rolling Deterministic Optimization (RDO)}
\label{RDO}
\begin{algorithmic}[1]
\STATE Initialize with $b=1$, $P^{past}_b =\emptyset$, $\tilde{x} = \emptyset$.
\FOR{$b = 1$ \TO $B$}
\vspace{2mm} 
\STATE \textbf{1. Information Update.} Patient arrival $P_b$ is realized between $s_{b-1}$ and $s_b-1$; obtain LOS estimates, $\{l_p: p\in P_b\}$.
\FOR{$p \in P^{past}_b$}
  \STATE  Update LOS estimate $l_p$ using the \textit{Information Update Procedure.}
\ENDFOR
\STATE \textbf{2. Schedule Optimization.} Solve deterministic BOP with $s_b$, $P_b$, $P^{past}_b$, $\tilde{x}$, $l_p$ $\forall p\in P_b\cup P^{past}_b$; implement the optimal solution $x^*$ for all $p\in P_b$.
\STATE \textbf{3. Schedule and Capacity Update.} 
\STATE $P_{b+1}^{past} \leftarrow P_b^{past} \cup P_b$
\STATE $\tilde{x}\leftarrow \tilde{x}\cup x^*$
\ENDFOR
\end{algorithmic}
\end{algorithm}

In practice, the sequence of $s_b$ is an additional design choice that can be set based on hospital practices and nature of different procedures. In all our experiments (both the deterministic formulation and stochastic or robust formulations in later sections), we set scheduling days to be the start of every month, and schedule patients in monthly batches given our context. For complex, elective cardiovascular procedures, there is usually a significant lead time between patient referral and when a patient undergoes surgery. For our patient cohort, the median lead time is 59 days and the average lead time is 79 days. Therefore, monthly scheduling is chosen as an example and proof of concept given the context: although patients need to wait for being scheduled, the effect of this additional wait time is limited in this case as the total wait time is being minimized under our optimization algorithms. In Appendix~\ref{app:additional_numerical_results}, we also re-run our numerical experiments for biweekly scheduling days and show that the resultant insights are similar. For other types of procedures that require faster scheduling, $s_b$ can be adjusted accordingly such that scheduling can happen more frequently in smaller batches. We discuss the implications of this design further in Section~\ref{Insights}.

\subsection{Model Performance under Perfect Information}\label{section:perfect_info}

Before introducing machine learning approaches for estimation of LOS, we first evaluate the performance of our models in the real world setting of the PAMC, using true historical LOS as model inputs, $\{l_p\}$. The purpose of this exercise is to identify the upper bound of the achievable level of operational performance versus the status quo, which sets the optimal performance benchmark for remaining numerical experiments.

\begin{figure*}
        \centering
        \begin{subfigure}{.5\textwidth}
        \centering
         \includegraphics[width=1\linewidth]{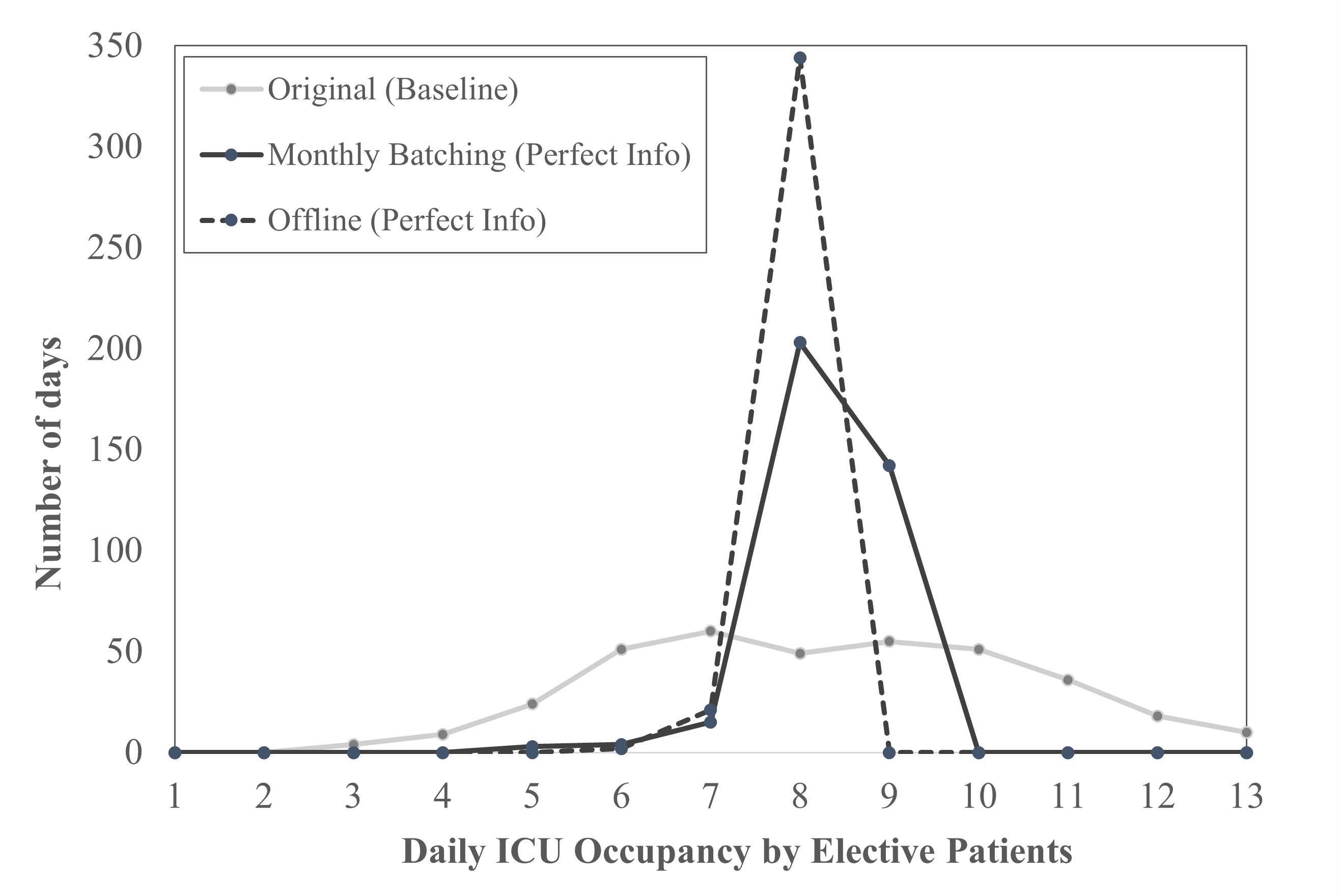}         
         \caption{Daily ICU occupancy under offline and monthly optimization}
        \label{fig:perfect_info_dist}
        \end{subfigure}%
        \begin{subfigure}{.5\textwidth}
        \centering
         \includegraphics[width=0.875\linewidth]{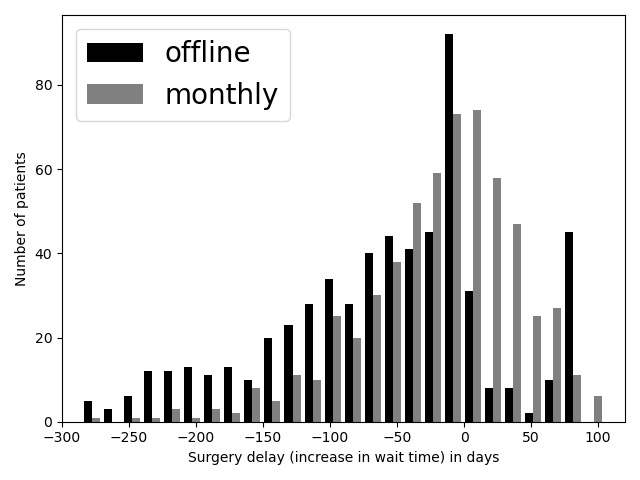}         
         \caption{Change in wait time versus the status quo}
         \label{fig:perfect_info_time}
    \end{subfigure}
    \caption{Schedule optimization eliminates days with 10 or more elective patients in the ICU given perfect information in LOS (left). Both offline and rolling monthly scheduling schedule the majority of the patients no later than their original surgery date (right). }
    \end{figure*}
    
We evaluate the performance of our scheduling model by considering the 596 elective cardiovascular surgeries that were carried out at the PAMC from September 2018 to March 2020. Patient arrivals are modeled using deterministic, historical arrivals. An optimal schedule was generated using the optimization models given the true LOS of each patient, and the resultant daily CVICU census was simulated based on the optimal surgical schedule and the actual LOS.

The performance of the optimization models is assessed using two metrics: 
1) percentage reduction in the number of high ICU occupancy days (10 or more elective patients) in the ICU versus the historical status quo, and 2) median and average reduction in patient wait times compared to the original schedule. Good performance corresponds to a measurable reduction in the number of high ICU occupancy days without increasing patient wait times. 

 When evaluating our models, we treat the first six months of the time period (Sept 2018 - March 2019) as a warm-up period\footnote{ 
 86 out of the 596 surgeries performed from September 2018 to March 2020 are associated with patient arrivals prior to September 2018. For these patients, the earliest possible surgery dates are adjusted so that their procedures are always scheduled on or after September 1st 2018. This modified constraint is an artifact of the fixed optimization time window and is tighter than what is realistically feasible. The first six months is treated as the warm up period for this reason.} and focus on model performance in the one-year period from March 2019 to March 2020. 

The performance of offline and rolling deterministic optimization is visualized in Figure \ref{fig:perfect_info_dist} and \ref{fig:perfect_info_time}. 
In both figures, we use $\beta=100$ for offline deterministic optimization and $\beta=25$ for rolling deterministic optimization.  
With perfect information on LOS, both models are able to eliminate the number of days with 10 or more elective patients in the ICU (Figure \ref{fig:perfect_info_dist}). Meanwhile, the offline model schedules 83.8\% of the patients no later than the status quo (Figure \ref{fig:perfect_info_time}), and the median and average reductions in wait time are 51.0 days and 66.1 days respectively. Rolling scheduling in monthly batches 
results in slightly longer wait times as the patients are not scheduled until the start of the following month. Still, 60.4\% of the patients are scheduled no later than the statues quo (Figure \ref{fig:perfect_info_time})), and the median and average reductions in wait time are 10.0 and 21.94 days respectively. Overall, both models achieve significant reduction in both high occupancy days in the ICU and patient wait times. 

The above results show that it is theoretically possible to eliminate the number of days with more than 10 elective patients in the ICU given perfect information on LOS at the time of scheduling. Such performance improvement can be achieved along with a reduction in average and median wait times. 

Having established the performance upper bound under perfect information, we next proceed to consider incorporating machine learning predictions of LOS in schedule optimization, and the impact of inaccurate predictions on deterministic optimization models.

\section{Predicting LOS with Machine Learning}\label{sec:los_predict}
 
Having established the performance upper bound of schedule optimization using accurate LOS predictions in the previous section, we now present the development of machine learning models for LOS prediction in this section, as well as the schedule optimization results using the predicted LOS.
 
We discuss both regression and classification models for LOS prediction.  5-fold cross validation is used for hyper-parameter tuning and model selection during the training process. The model with the highest cross-validation score is selected for validation on the test set. The sklearn library \cite{pedregosa2011scikit} in Python was used to train and implement the selected models.

The predicted LOS is then used as input for RDO. Scheduling optimization is only performed on the test set and not on the training set. This simulates the real world setting where the incoming patients are prospectively scheduled based on LOS predictions at the time of admission. We compare the schedule optimization result with the status quo.

\subsection{The Dataset}
Our data set consists of medical records, surgery data and CVICU census data for 2,352 elective cardiovascular surgical patients at the PAMC from 2014 to 2020. It is split into a training set with 1,738 patients from January 2014 to mid-June 2018, and a test set with 614 surgical patients from mid-June 2018 to May 2020. 

The patient features are chosen based on clinical knowledge of the experienced surgeons at the unit, as detailed below.\\

\textbf{Categorical Features:}
\begin{itemize}[label=$\bullet$]
    \item Surgeon: The assigned surgeon for the surgery
    \item Procedure: The type of surgical procedure 
    \item Month: The month when the surgery is performed
    \item Genetic Disorder: The type of genetic disorder the patient has prior to the surgery, if any 
    \item Ventilator Status: whether the patient is on a ventilator prior to the surgery
    \item Respiratory Status: the type of respiratory diseases the patient has, if any, prior to the surgery
\end{itemize}

\textbf{Continuous Features:}
\begin{itemize}[label=$\bullet$]
\item Age: The age of the patient at the time of the surgery
\item Weight: The weight of the patient at the time of surgery
\item Height: The height of the patient at the time of surgery
\end{itemize}

The missing values in the data are imputed using the mean for continuous features and the mode for categorical features from the training set.

To compare our data quality with the existing literature, we first train a binary classification model to identify patients with risks of prolonged stays of more than 5 or 10 days, making up 28\% and 12.8\% of the training set respectively. Candidate
machine learning models considered include logistic regression, random forest and gradient boosting machine. Gradient boosting machine is selected based on cross-validation AUC.

We refer to Ettema et al. (2010)\cite{ReviewEttema10} for performance benchmarks. The authors of \cite{ReviewEttema10} reviewed and validated 14 models for the prediction of prolonged LOS in the ICU after cardiac surgery using data from 11,395 surgeries, and found that the area under the curve (AUC) scores of the best performing models range from 0.71 to 0.75. Our model achieves an AUC score of 0.77 for stays more than 5 days and 0.73 for stays more than 10 days on the test set, which is inline with that of the best-performing models identified in \cite{ReviewEttema10} (See Table~\ref{table:auc}).

\begin{table}[!h]
\centering
\setlength{\tabcolsep}{8pt}
\setcellgapes{4pt}
\makegapedcells
\begin{tabular}{cc}
\hline
\textbf{Model} & \textbf{AUC Score }\\ \hline
$>$5 days             & 0.77              \\ \hline
$>$10 days            & 0.73              \\ \hline
Benchmarks            & 0.71-0.75\cite{ReviewEttema10}\\ \hline
\end{tabular}
\caption{Model performance in binary classification vs benchmarks}
\label{table:auc}
\end{table}

Although a direct comparison with existing literature is difficult due to different patient populations and different definitions of prolonged stay, our model and features can achieve high levels of distinguishing performance compared to the benchmarks.

\subsection{Model Development}\label{sec:tradeoff}

We now train machine learning models for LOS with the objective of using the predictions for surgical scheduling.

We first consider classification models that classify LOS into ordinal buckets: 0-1 day, 2-5 days, 6-10 days and $>10$ days. For each LOS bucket, we define two metrics to evaluate model performance. `Accuracy 1' is calculated as the exact number of true predictions in the bucket divided by the actual number of patients in the bucket. `Accuracy 2' extends the definition of true prediction to include the scenario where the patient is predicted to be in the LOS bucket adjacent to his or her true bucket. 

Candidate models include gradient boosting machine (GBM), random forest and ordinal regression models. GBM achieves the best cross-validated accuracy score on LOS classification with multiple LOS buckets. The model achieves an overall accuracy of 53\%, and 88\% of the predicted LOS either fall in the true bucket or the adjacent buckets. The most significant predictors according to the impurity-based feature importance are procedure types (0.47), weight (0.22), height (0.17) and the surgeon (0.07). 

A breakdown of the predictive accuracy on each buckets of the test set is presented in Table~\ref{table:prediction}. While predictive accuracy for the first three patient groups with LOS $\leq$ 10 days is relatively high based on Accuracy 2, predictive accuracy drops sharply for the group of patients with longer than 10 days of LOS, where the majority of the patients' LOS is being under-estimated. Similar behavior is observed under other choices of bucketing, such as using finer buckets for the LOS$>10$ patient group.

\begin{table}[!h]
\setlength{\tabcolsep}{2pt}
\setcellgapes{4pt}
\makegapedcells
\begin{tabular}{ccc}
\hline
\textbf{Bucket (\% of Test Set)} & \textbf{Accuracy 1} & \textbf{Accuracy 2} \\ \hline
0-1 days (26\%)                   & 0.33                & 0.96                \\ \hline
2-5 days (41\%)                   & 0.89               & 0.97               \\ \hline
6-10 days (15\%)                  & 0.11                & 1.00              \\ \hline
$>10$ days (18\%)                  & 0.33                  & 0.43                 \\ \hline
\textbf{Overall  }              &\textbf{0.53}            & \textbf{0.88}  \\ \hline
\end{tabular}
\caption{Model performance (GBM) on LOS classification using the coarse LOS buckets.}
\label{table:prediction}
\end{table}

\begin{figure*}[!h]
    \centering
    \includegraphics[width=0.95\linewidth]{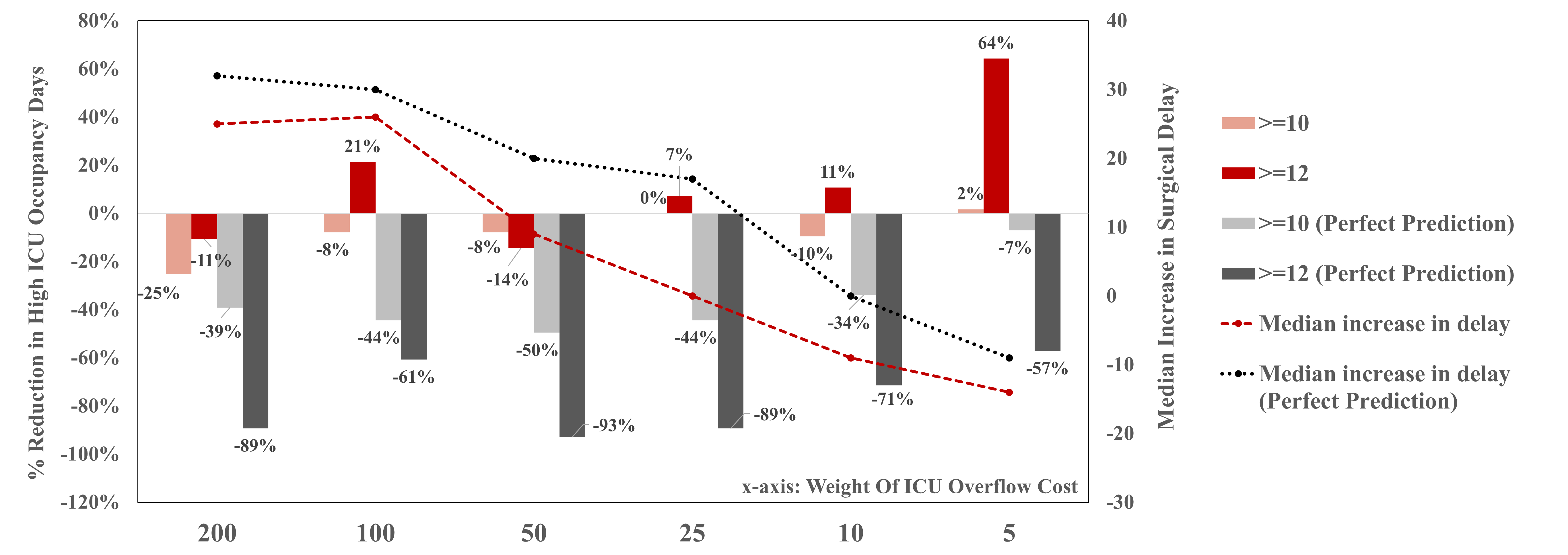}    
    \caption{RDO performance comparison using 100\% accurate classification predictions (grey) and actual machine learning predictions (red).}
    \label{fig:opt_with_prediction}
\end{figure*}

We use the predicted buckets of LOS are as inputs for RDO in Algorithm~\ref{RDO}. Given each patient's predicted LOS bucket, the upper bound of the first three buckets is used as the LOS point estimates $l_p$, and $l_p=30$ is used for the $>10$ bucket because 95\% of the patients left the CVICU in less than or equal to 30 days. Although the machine learning model uses month of surgery from historical data, this feature was studied and had little effect on LOS prediction.
Optimization performance on ICU occupancy and wait times is shown in red in Figure~\ref{fig:opt_with_prediction}, in comparison with performance upper bounds in grey assuming the classification algorithm achieves 100\% accuracy. Here, we consider a range of weights on the cost of ICU overflow relative to the cost of total wait times: $\beta \in \{200,100,50,25,10,5\}$. With perfect predictions of each patient's LOS bucket (grey bars), the optimization model significantly reduces the number of high ICU occupancy days without pushing back surgeries when $\beta = 10$ and 5. In contrast, the machine learning predicted outcomes show only slight improvement in both objectives for $\beta=50$ (red bars). The trade off between the two objectives becomes significant as $\beta$ decreases. When the weight on the cost of ICU overflow is sufficiently high, optimization reduces the number of high ICU occupancy days at the cost of increasing patient wait times and vice versa. 

\begin{figure}[!h]
    \centering
    \includegraphics[width=1\linewidth]{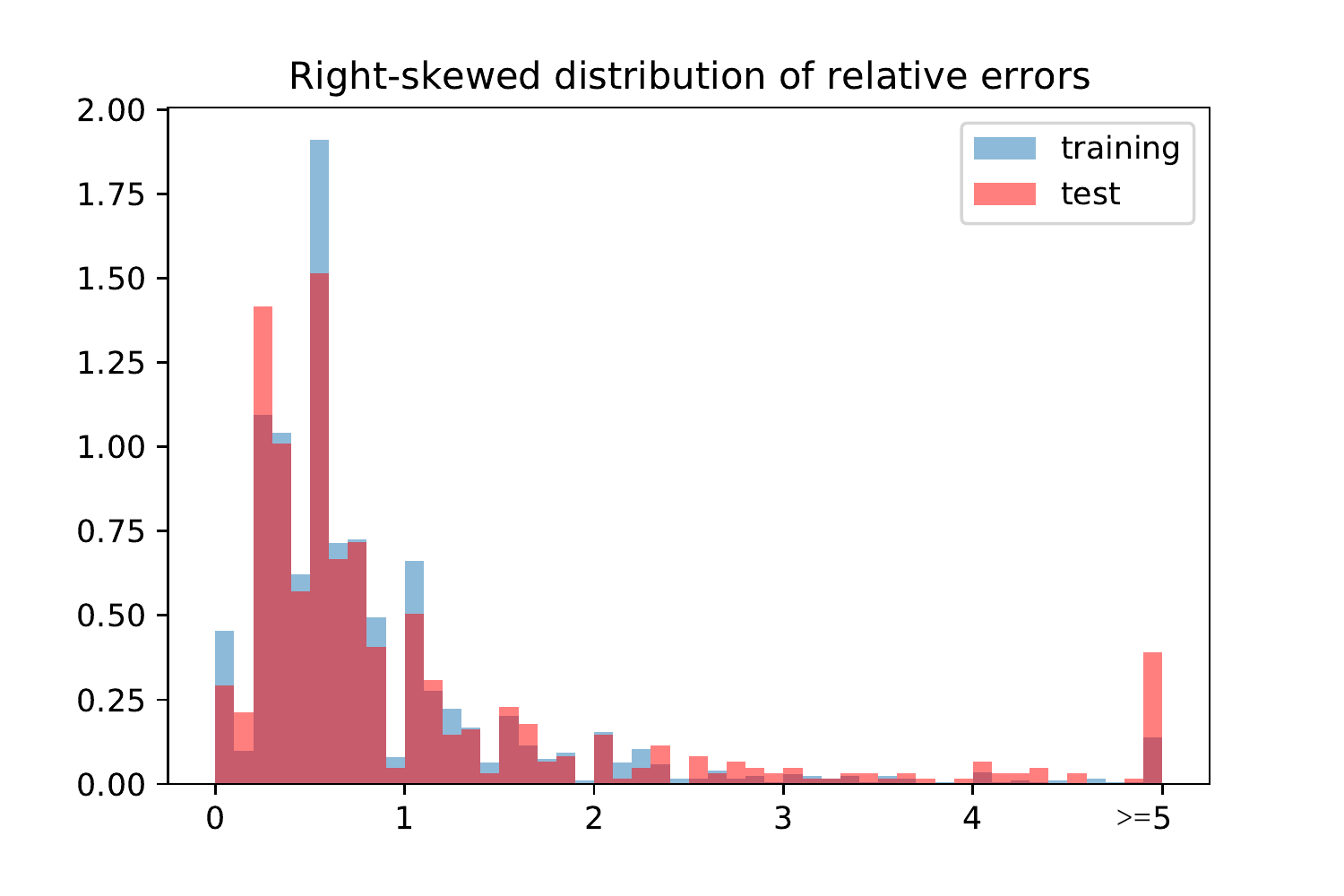}
    \caption{Relative predictive errors of the GBM regression model, $\frac{\text{true LOS}}{\text{predicted LOS}}$.}
    \label{fig:err_dist}
\end{figure}

Compared to classification models, generating point-forecast for individual patients using regression models is even more challenging especially for those with prolonged LOS (see, for example, \cite{Tsai16}). We follow the same procedure to select and train a variety of regression models including OLS regression, Lasso regression, gradient boosting machine, quantile regression and more. However, when the generated point-predictions are combined with deterministic optimization, there is still no model that achieves performance improvement on both metrics of interest. Similar to the classification model above, the main difficulty comes from significant under-estimation of prolonged LOS. Figure~\ref{fig:err_dist} plots the density distribution of relative predictive errors, defined as the ratio between true LOS and predicted LOS, for training and testing set. The distribution is heavily right-skewed with a long-tail of very large relative errors (i.e. true LOS $>>$ predicted LOS). We explore this issue further in the following section.

\subsection{Identifying the Challenges
 }

   \begin{figure*}[!h]
        \centering
        \begin{subfigure}{.5\textwidth}
      \centering
    \includegraphics[width = 1\linewidth]{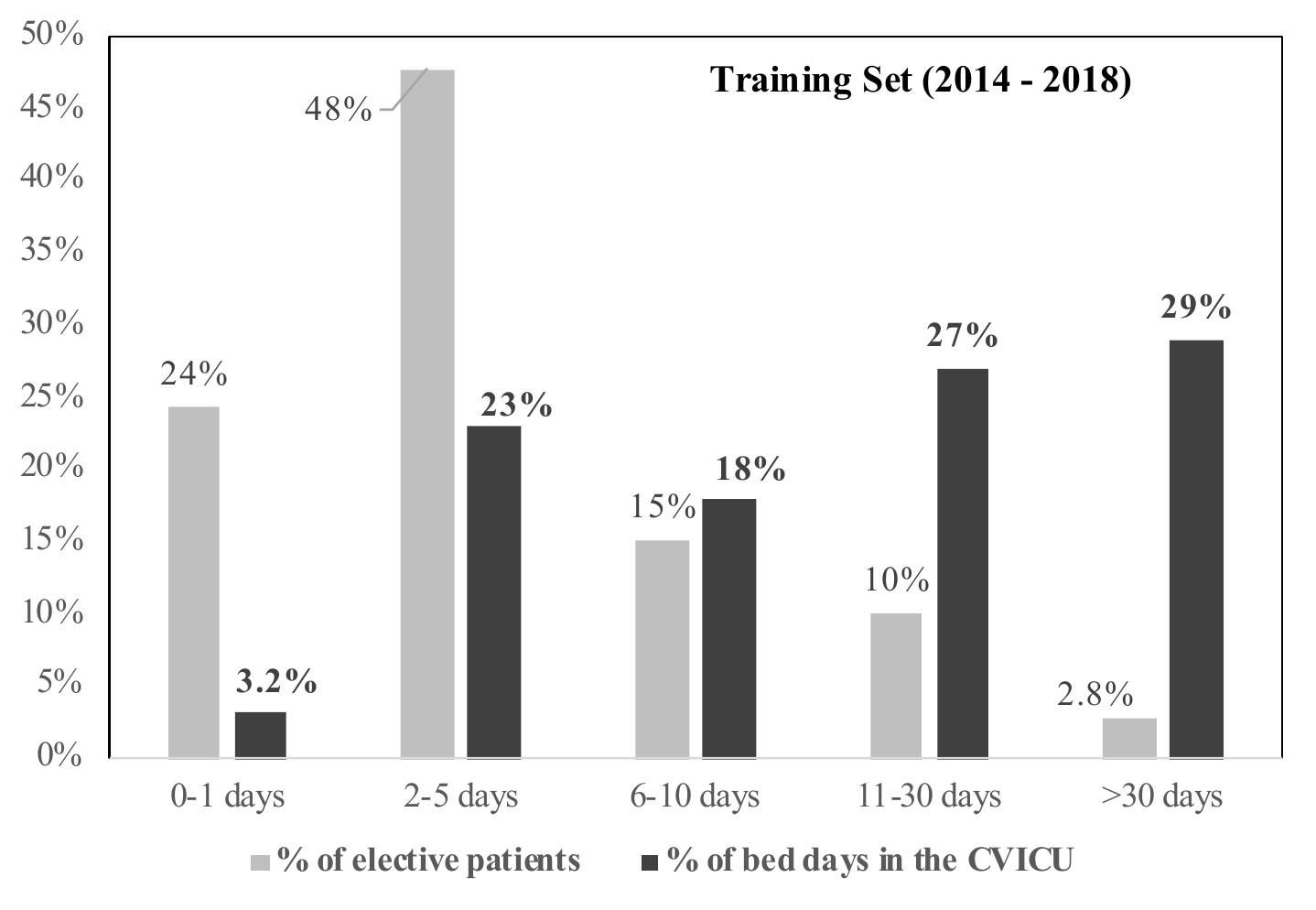}
    \caption{Training set}
    \label{fig:nonstationary}
        \end{subfigure}%
        \begin{subfigure}{.5\textwidth}
     \centering
    \includegraphics[width = 1\linewidth]{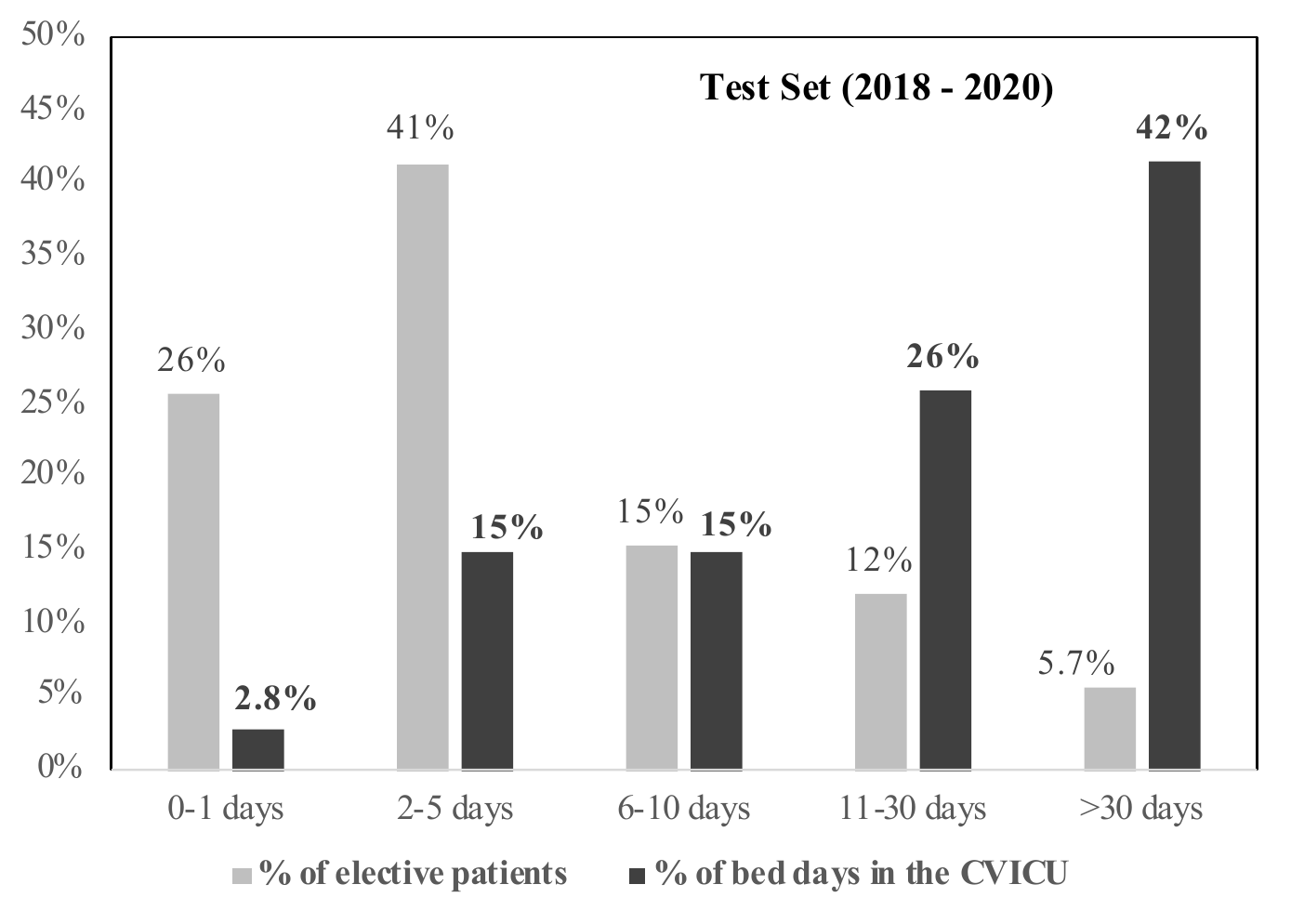}
    \caption{Test set}
    \label{fig:long_tail}
    \end{subfigure}
    \caption{The fraction of patients with extended LOS ($>30$ days) doubled in the test set (right) compared to the train set (left). 5.7\% of the patients with LOS of $>30$ days make up 42\% of the CVICU census count from 2018 to 2020. Training set is from January 2014 to mid-June 2018. Test set is from mid-June 2018 to May 2020. }
      \label{fig}
    \end{figure*}

Results in Section \ref{sec:tradeoff} show that the major challenge in LOS prediction and achieving measurable improvements through optimization is predicting prolonged LOS at the time of admission. This is consistent with what many past studies have observed. For example, neither linear regression or artificial neural network developed in \cite{Tsai16} is able to predict patient LOS of greater than 15 days. The neural network proposed in \cite{lafaro2015neural} is unable to predict patient LOS of greater than 150 hours on the validation set. \cite{yang2010predicting} concluded that the predictive accuracy of most existing prediction techniques is expected to be inferior in the tail of the distribution to that in the middle due to imbalanced data. Furthermore, \cite{kramer2010predictive} shows that after 5 days in the ICU, the most significant predictors of the remaining LOS are features collected on day 5 instead of those at the time of admission.

Poor predictions of prolonged LOS have a significant impact on optimization performance, especially in the context of cardiovascular surgeries where LOS in the ICU tends to have a long tail distribution. As shown in Figure~\ref{fig:long_tail}, 5.7\% of the patients with LOS of greater than 30 days in the ICU makes up 42\% of the total CVICU daily census count in the test set aggregated from 2018 to 2020. This small group of patients thus has a disproportionately large influence on ICU resource use and operational performance.

A closer examination of our data also reveals the non-stationary nature of hospital operations, another challenge to developing and evaluating predictive models in real life. A comparison between Figure~\ref{fig:long_tail} and \ref{fig:nonstationary} highlights the shift in patient population in terms of LOS distributions from 2014-2018 to 2018-2020 at the PAMC. The most notable change is an increase in cases with extended LOS in the CVICU, driven by the hospital's decision to admit a larger fraction of complex cases in recent years. Such shifts in hospital operations make prospective predictions of LOS based on past data even less reliable. We split the training and test set in chronological order instead of randomly to reflect the impact of such inherent non-stationarity on model development and performance in practice.

\section{Data-Driven Surgical Scheduling with Machine Learning Under Uncertain LOS}\label{sec:sp}

While combining machine learning with a deterministic formulation of optimization can provide significant performance improvement in theory, we have shown that the unpredictability of the long-tailed LOS along with other challenges can lead to poor performance in practice. A natural next step is to consider optimization methods that directly incorporate the uncertainties in LOS.

In this section, we propose a data-driven surgical scheduling framework, designed to address the identified challenges in the previous sections. Our framework combines optimization under uncertainty with machine-learning predicted distribution of LOS, and rolling information update.

Under this general framework, we develop three algorithms: Standard-RSO, Conservative-RSO and RRO. Standard-RSO and Conservative-RSO apply stochastic optimization using the predicted LOS distributions to minimize the expected cost, where the latter further adjusts the predicted LOS distribution to target under-estimations for prolonged LOS. RRO uses robust optimization to produce solutions that are robust against under-estimations for prolonged LOS.

\subsection{Estimating LOS Probability Distribution using Machine Learning}

Instead of relying on deterministic predictions of LOS for optimization, we now use machine learning models to predict the probability distribution of LOS. We start by generating point-wise prediction of LOS for each patient using conventional machine learning models discussed in Section \ref{sec:los_predict}. For each patient $p$, let $v_p$ be the value of LOS point prediction from the machine learning model for the true LOS, $L_p$, and  define the relative prediction error 
\begin{gather}\label{eqn:error-standard}
 R_p = L_p / v_p.   
\end{gather}
$L_p$ and $R_p$ are random variables and $v_p$ is known. 
We assume that the random variables $R_p$'s are independent and identically distributed across patients from a distributions $\mathcal{S}$, i.e., 
$$R_p\overset{\text{iid}}{\sim} \mathcal{S}\quad \forall p\in P.$$
Note that $R_p>1$ represents an under-prediction of LOS and $R_p<1$ means an over-prediction. We choose to model \textit{relative} prediction errors so that the absolute prediction errors, $|L_p-v_p|$, tend to be larger for patients with prolonged realized LOS. This is in accordance with our finding in Section \ref{sec:los_predict} that prediction is more challenging for prolonged LOS.

The distribution $\mathcal{S}$ is approximated using the empirical distribution of predictive errors of the machine learning model during training and validation. Figure~\ref{fig:err_dist} provides an example of this distribution. 

Given estimated $\mathcal{S}$, the distribution of $L_p$ for individual patient $p$ is estimated using
$$L_p = round(v_p\cdot R_p)\quad R_p\sim \hat{\mathcal{S}}.$$

\subsection{Rolling Stochastic Optimization (RSO)}\label{sec:rsp}

Building on the rolling deterministic optimization framework introduced in section~\ref{sec:batch}, we first explore stochastic optimization approaches that utilize estimated probability distributions of patient LOS. 

Let the random variable $L_p$ denote the LOS of each patient $p$, and a \textit{LOS realization trace}, $$\omega = \{l_p: p \in P_b \cup P^{past}_b\},$$ denote a sequence of realized LOS for all patients who have arrived at the system at the start of period $b$. We use $\mu_b$ to denote the discrete probability distribution of $\omega$ over the set of all possible traces for batch $b$, denoted as $\Omega_b$. We assume that each patient's LOS is independent of other patients, and hence $\mu_b$ is the joint distribution of independent random variables $L_p$, for all $p\in P_b\cup P_b^{past}$.

Using the same set of notations as deterministic BOP introduced in Section~\ref{sec:batch}, stochastic optimization formulations aim to minimize the expected cost taken over the distribution $\mu_b$:
\begin{equation}
    \min_{x}\,\,\sum_{p\in P_b}
    \sum_{d=s_b}^{N_d}(d - d^{min}_p)^+ x_{d,p} + \beta \cdot \mathbb{E}_{\omega\sim \mu_b} \sum_{d=s_b}^{N_d} f(u_d(x,\omega)) \label{sp:objective}
\end{equation}
Equation~\eqref{sp:objective} parallels the deterministic objective function of deterministic BOP in Equation~\eqref{opt2:objective}. The distinction is that the ICU overflow variable, $u_d$ is now a random variable that is a function of $x$ and $\omega$.

Following the approach used in \cite{min_yih}, we approximate the objective function \eqref{sp:objective} using Sample Average Approximation (SAA). Specifically, we randomly sample $N_{\Omega}$ traces of $\omega$ from $\Omega_b$, according to probability distribution $\mu_b$. Each trace of $\omega$ is sampled by independently sampling $l_p$ from the estimated distribution of $L_p$ for each patient, $p\in P_b\cup P_b^{past}$. 

Denote sampled traces as $\{\omega^{(n)}: n=1,\ldots,N_{\Omega}\}$ and sampled LOS $\{l^{(n)}_p:n=1,\ldots,N_{\Omega},p\in P_b\cup P^{past}_b\}$. For each sampled trace, we introduce auxiliary binary variables $y^{(n)}_{d,p}$ that indicate if patient $p$ needs an ICU bed on day $d$ given $l^{(n)}_p$. Similarly, we use $u^{(n)}_{d}$ to denote ICU overflow on day $d$ given $\omega^{(n)}$. The stochastic BOP with SAA is formulated below.
\begin{subequations}\label{eqns:saa}
\begin{equation}
    \min_{x}\, \sum_{p\in P_b}
    \sum_{d=s_b}^{N_d}(d - d^{min}_p)^+ x_{d,p} +  \frac{\beta}{N_\Omega}\sum_{n=1}^{N_\Omega}\sum_{d=s_b}^{N_d} f(u^{(n)}_d) \label{sp:saa}
\end{equation}
\begin{equation}
x_{d,p} = \tilde{x}_{d,p}\quad \forall p\in P^{past}_b, d\in D
\label{spcon:past}
\end{equation}
\begin{equation}
    \sum_{d=s_b}^{N_d} x_{d,p} = 1,\quad \sum_{d=1}^{s_b -1} x_{d,p}=0 \quad \forall p\in P_b \label{spcon:match}
\end{equation}
\begin{equation}
\begin{aligned}
y^{(n)}_{d,p} = \sum_{d'=\max(d-l^{(n)}_p+1,1)}^d x_{d',p} &\quad \forall p\in P_b\cup P_b^{past}, \\
& \forall d\in D, n=1,\ldots, N_{\Omega}
\label{spcon:y}
\end{aligned}
\end{equation}
\begin{equation}
    \sum_{p\in P_b \cup P^{past}_b} y^{(n)}_{d,p} \leq c + u^{(n)}_d \quad \forall d\in D, n=1,\ldots,N_\Omega \label{spcon:overflow}
\end{equation}
\begin{equation}
x\in Q^{op}  \label{spcon:operations}
\end{equation}
\begin{equation}
   y^{(n)}_{d,p},x_{d,p}\in\{0,1\},\quad u_d\geq 0\label{spcon:vars}
\end{equation}
\end{subequations}
The objective function~\eqref{sp:saa} formulates the SAA approximation of Equation~\eqref{sp:objective}. Constraints \eqref{spcon:y} and \eqref{spcon:overflow} formulate the stochastic parallel to constraints \eqref{con:y2} and \eqref{con:overflow2} in deterministic BOP, creating a replica of variables $y_{d,p}$ and $u_{d,p}$ for every sample trace $\omega^{(n)}$. The other constraints remain identical to their deterministic counterpart. Note that we focus on surgical scheduling problems where different realizations of $\omega$ only affect daily ICU occupancy and overflow, and do not affect feasibility of a schedule, $x$. In practice, ICU overflow can be accommodated by setting up temporary ICU beds or utilizing spare resources from other ICUs. Feasibility is thus determined exclusively by operational constraints such as OR room capacity, surgeon and patient availability, which remain unchanged.

We combine stochastic BOP with the rolling optimization framework introduced previously for RDO. Under this framework, stochastic BOP is solved sequentially for a sequence of schedule days, $\{s_b: b = 1,\ldots,B\}$. Meanwhile, the probability distributions of LOS $L_p$ are progressively updated for every patients as uncertainty realizes. We introduce the information update procedure for stochastic BOP below.

\paragraph{Information Update Procedure for Stochastic BOP.} At the start of each batch $b$, the distribution of $L_p$ for all $p \in P^{past}_b$ is updated as follows.
    \begin{enumerate}[label=(\alph*)]
        \item  If the patient has undergone surgery and has been discharged from ICU by scheduling day $s_b$, her realized LOS is observed and we update $L_p$ to be a constant equal to the true LOS.
        \item If the patient is in the ICU on day $s_b$ having stayed for $m$ days, then we update the distribution of $L_p$ using the conditional distribution of $L_p|L_p\geq m$. To obtain the resultant distribution, we update the distribution of prediction errors, $R_p\sim \mathcal{S}$ by conditioning on
        $round(v_p\cdot R_p)\geq m$ while fixing $v_p$.
    \item If the procedure of the patient is scheduled after $s_b$, there is no change to the distribution of $L_p$.
    \end{enumerate}

We develop two rolling stochastic optimization algorithms, Standard-RSO and Conservative-RSO. Both algorithms adopt the same framework combining stochastic optimization, machine-learning predicted distribution and rolling information update.

The two algorithms differ from each other on how LOS realization traces $\omega$ are sampled for SAA. In Standard-RSO, realizations of $l^{(n)}_p$ are generated by sampling the relative prediction error $r^{(n)}_p$ directly from $\mathcal{S}$ and setting $$l^{(n)}_p = round(v_p\cdot r^{(n)}_p).$$ In contrast, in Conservative-RSO, we generate \textit{conservative} realizations of $l^{(n)}_p$ by rounding any $r^{(n)}_p<1$ drawn from $\mathcal{S}$ to one, i.e., $$l^{(n)}_p = round(v_p\cdot \max\{r^{(n)}_p,1\}).$$
In other words, Standard-RSO draws LOS realizations from a distribution derived from two-way predictive errors, $\mathcal{S}$; Conservative-RSO draws LOS realizations from an adjusted distribution of $\mathcal{S}$, where only one-way errors from LOS \textit{under-predictions} (i.e. $r_p>1$) are retained, and any over-prediction errors $r_p<1$ are rounded up to 1. Conservative-RSO is designed such that the algorithm focuses on stochasticity arising from difficult-to-predict extended LOS for cardiovascular surgeries, where machine learning models consistently yield under-predictions. 

We formalize the two rolling stochastic optimization algorithms in Algorithms~\ref{SRSO} and \ref{CRSO} below.

\begin{algorithm}
\caption{Standard-RSO}
\label{SRSO}
\begin{algorithmic}[1]
\STATE Initialize with $b=1$, $P^{past}_b =\emptyset$, $\tilde{x} = \emptyset$.
\FOR{$b = 1$ \TO $B$}
\vspace{2mm} 
\STATE \textbf{1. Information Update.} Patient arrival $P_b$ is realized between $s_{b-1}$ and $s_b-1$; obtain point-prediction of LOS, $\{v_p: p\in P_b\}$.
\FOR{$p \in P^{past}_b$}
\STATE Update LOS distribution $L_p$ and $\mathcal{S}$ using the \textit{Information Update Procedure} for stochastic BOP.
\ENDFOR
\STATE \textbf{2. SAA Sampling}
\FOR{$n=1,2,\ldots, N_\Omega$}
\STATE Sample the LOS realization trace, $\omega^{(n)} =  \{l^{(n)}_p:p\in P_b\cup P_b^{past}\}$ by randomly sampling relative prediction error $r^{(n)}_p$ from $\mathcal{S}$ and setting $l^{(n)}_p = round(r^{(n)}_p \cdot v_p)$ for all $p$. In the case where $L_p=l_p$ is a constant, set $l^{(n)}_p = l_p$. 
\ENDFOR
\STATE \textbf{3. Schedule Optimization.} Solve stochastic BOP with SAA given $s_b$, $P_b$, $P^{past}_b$, $\tilde{x}$, $\{\omega^{(n)}$: $n =1,2,\ldots, N_\Omega$\}; implement the optimal solution $x^*$ for all $p\in P_b$.
\STATE \textbf{4. Schedule and Capacity Update.} 
\STATE $P_{b+1}^{past} \leftarrow P_b^{past} \cup P_b$
\STATE $\tilde{x}\leftarrow \tilde{x}\cup x^*$
\ENDFOR
\end{algorithmic}
\end{algorithm}

\begin{algorithm}
\caption{Conservative-RSO}
\label{CRSO}
\begin{algorithmic}[1]
\STATE Initialize with $b=1$, $P^{past}_b =\emptyset$, $\tilde{x} = \emptyset$.
\FOR{$b = 1$ \TO $B$}
\vspace{2mm} 
\STATE \textbf{1. Information Update.} Patient arrival $P_b$ is realized between $s_{b-1}$ and $s_b-1$; obtain point-prediction of LOS, $\{v_p: p\in P_b\}$.
\FOR{$p \in P^{past}_b$}
\STATE Update LOS distribution $L_p$ and $\mathcal{S}$ using the \textit{Information Update Procedure} for stochastic BOP.
\ENDFOR
\STATE \textbf{2. Conservative SAA Sampling}
\FOR{$n=1,2,\ldots, N_\Omega$}
\STATE Sample the LOS realization trace, $\omega^{(n)} =  \{l^{(n)}_p:p\in P_b\cup P_b^{past}\}$ by randomly sampling relative prediction error $r^{(n)}_p$ from $\mathcal{S}$ and setting $l^{(n)}_p = round(v_p \cdot \max\{r^{(n)}_p,1\} )$ for all $p$. In the case where $L_p=l_p$ is a constant, set $l^{(n)}_p = l_p$. 
\ENDFOR
\STATE \textbf{3. Schedule Optimization.} Solve stochastic BOP with SAA given $s_b$, $P_b$, $P^{past}_b$, $\tilde{x}$, $\{\omega^{(n)}$: $n =1,2,\ldots, N_\Omega$\}; implement the optimal solution $x^*$ for all $p\in P_b$.
\STATE \textbf{4. Schedule and Capacity Update.} 
\STATE $P_{b+1}^{past} \leftarrow P_b^{past} \cup P_b$
\STATE $\tilde{x}\leftarrow \tilde{x}\cup x^*$
\ENDFOR
\end{algorithmic}
\end{algorithm}

When implementing Standard-RSO and Conservative-RSO, we set $N_{\Omega} = 10$ to obtain our numerical results. Large values of $N_{\Omega}$ can lead to very large numbers of variables and constraints under Equations~\eqref{spcon:y} and \eqref{spcon:overflow}. We thus limit the value of $N_{\Omega}$ due to computational constraints. We discuss the computational limitations further in Section~\ref{Insights}.

\subsection{Rolling Robust Optimization (RRO)}\label{sec:rro}
Robust optimization formulations provide an alternative to stochastic optimization for scheduling under uncertainty. Robust optimization aims to minimize the \textit{worst-case} cost defined over an uncertainty set, $\mathcal{U}_b$, which is a chosen subset of all possible realized traces $\omega = \{l_p:p\in P_b\cup P^{past}_b\}$. Using the same notations as before, the formulation of robust BOP for a pre-defined uncertainty set $\mathcal{U}_b$ can be written as follows.
\begin{subequations}\label{eqns:brop}
\begin{equation}
    \min_{x}\quad \sum_{p\in P_b}
    \sum_{d=s_b}^{N_d}(d - d^{min}_p)^+ x_{d,p} + \beta \cdot \theta(x)
 \label{eqn:ro-obj}
\end{equation}
 s.t.
\begin{equation}
x_{d,p} = \tilde{x}_{d,p}\quad \forall p\in P^{past}_b, d\in D
\label{rocon:past}
\end{equation}
\begin{equation}
    \sum_{d=s_b}^{N_d} x_{d,p} = 1,\quad \sum_{d=1}^{s_b -1} x_{d,p}=0 \quad \forall p\in P_b 
\end{equation}
\begin{equation}
x \in Q^{op}, x_{d,p}\in\{0,1\}
\end{equation}
where $\theta(x)$ is the worst-case cost associated with ICU overflow for all possible LOS realization traces in the uncertainty set,
\begin{equation}\label{brop-theta-start}
    \theta(x) = \max_{\{l_p: p\in P_b\cup P^{past}_b\} \in\mathcal{U}_b} \quad \min_{y,u} \sum^{N_d}_{d=s_b} f(u_d)
\end{equation}
s.t.
\begin{equation}
   \begin{aligned}
   y_{d,p} = \sum_{d'=\max(d-l_p+1,1)}^d x_{d',p} \,\, \forall p\in P_b\cup P^{past}_b, d\in D
   \end{aligned} 
   \label{rocon:y}
\end{equation}
   \begin{equation}
    \sum_{p\in P_b \cup P^{past}_b} y_{d,p} \leq c + u_d \quad \forall d\in D
\end{equation} 
\begin{equation}\label{brop-theta-end}
y _{d,p}\in\{0,1\}, u_d\geq 0.
\end{equation}
\end{subequations}
Apart from the uncertainty set $\mathcal{U}_b$, the remaining constraints of robust BOP are similar to those of deterministic BOP in Equations~\eqref{eqns:deterministic BOP}. It is worth noting that in Equation~\eqref{rocon:y}, $x_{d,p}$ are now constant model parameters passed on from the outer minimization problem, and $l_p$ are now decision variables to be optimized.

To formulate the uncertainty set $\mathcal{U}_b$ for scheduling day $s_b$, let $l_p^{min}$ and $l_p^{max}$ denote the lower and upper bound of LOS for patient $p$. and $A_b$ the subset of patients where $l_p^{max} - l_p^{min}>0$, i.e., $A_b = \{ p\in P_b \cup P^{past}_b:  l^{max}_p - l_p^{min}>0\}$. The uncertainty set, $\mathcal{U}_b$ is defined to be the set of possible realizations of $\omega \in \Omega_b$ that satisfy the following two constraints:
\begin{equation}
    l_p^{min} \leq l_p\leq l_p^{max}\quad \forall p \in P_b \cup P^{past}_b,\label{con:u1}
\end{equation}
and, for some chosen constant $0\leq \eta\leq 1$,
\begin{equation}
  \sum_{p\in A_b}\left[ \frac{l_p - l_p^{min}}{l^{max}_p - l_p^{min}}\right]\leq \eta\cdot |A_b|. \label{con:u2}
\end{equation}
Constraint~\eqref{con:u2} enforces a \textit{budget of uncertainty}, $\eta\cdot |A_b|$, which limits the total possible normalized deviations (i.e., extended ICU days) from the lower bound $l^{min}_p$. Following the terminology in robust optimization literature (see, e.g., \cite{bertsimas2004price}), we refer to $\eta$ as the \textit{uncertainty budget}, and tune the value of $\eta$ among $\{0.5, 0.75, 1.0\}$ based on performance of numerical experiments.

Instead of using stylized assumptions to set values for $l_p^{min}$ and $l_p^{max}$ and construct the uncertainty set (see, e.g., \cite{robust_opt}), we tailor the values of $l_p^{min}$ and $l_p^{max}$ for individual patients using machine-learning predicted LOS distributions. For all incoming patients, $p\in P_b$, we set $l_p^{min}$ to be the point-prediction given by the machine learning model, $v_p$. The LOS upper-bound $l_p^{max}$, on the other hand, is determined using the $100(1-\alpha)$ percentile of relative predicted error under the estimated distribution, $\mathcal{S}$: $$l_p^{max} = round(v_p \cdot r_{max}),\quad r_{max} = F_{\mathcal{S}}^{-1}(1-\alpha),$$
where $F^{-1}_{S}$ denotes the inverse cumulative distribution function of $\mathcal{S}$. For our numerical experiments, we present results with $\alpha=0.2$ (i.e., the 80th percentile), chosen among $\{0.1, 0.15,0.2,0.25\}$ based on simulated robust optimization performance for our data set. 

As uncertainties of LOS realize overtime for previously scheduled patients, $P^{past}_b$, we update the distributions of LOS and parameters used for the uncertainty set dynamically as follows.
\paragraph{Information Update Procedure for Robust BOP.} At the start of each batch $b$, update the distributions of $L_p$ for all $p \in P^{past}_b$ using the update procedure for stochastic BOP introduced in Section~\ref{sec:sp}. Given the updated distributions, $l_p^{min}$, $l_p^{max}$ are updated as follows.
\begin{enumerate}[label=(\alph*)]
    \item If the patient has undergone surgery and has been discharged from ICU by scheduling day $s_b$, update $l_p^{min} = l_p^{max}$ to be equal to the true LOS.
    \item  If the patient is in the ICU on day $s_b$ having stayed for $m$ days, update
    $l_p^{min} = \max\{m,v_p\}$, and
    $l_p^{max} = round(v_p \cdot r_{max})$, where $r_{max} = F_{\mathcal{S}}^{-1}(1-\alpha)$ using the updated $\mathcal{S}$.
    \item If the procedure of the patient is scheduled after $s_b$, there is no change to $l_p^{min}$, $l_p^{max}$.
\end{enumerate}
We develop an adapted column and constraint generation approach (AC\&CG) to solve robust BOP in \eqref{eqns:brop} with the above definition of $\mathcal{U}_b$. The method of AC\&CG for surgical scheduling with ICU capacity constraints was first proposed in \cite{robust_opt}, which was designed for linear cost functions $f(u_d)$ only. Our AC\&CG approach extends the algorithm in \cite{robust_opt} to accommodate convex, piece-wise linear formulations of $f(u_d)$. 

At a high level, the AC\&CG approach involves iteratively solving a \textit{Main problem} and a \textit{Recourse problem}. Every iteration $t$, the Main problem solves for a \textit{temporary} optimal scheduling decision, $x^{t*}$, that minimizes the worst-case cost over a \textit{subset} of the uncertainty set, $\Omega^t \subset \mathcal{U}_b$. The optimal objective value provides a lower bound to that of the original BOP. The Recourse problem then finds a worst-case LOS realization trace, $\omega^{t*}\in \mathcal{U}_b$, that maximizes the cost of ICU overflow under the temporary scheduling decision, $x^{t*}$. The optimal objective value provides an upper bound to that of the original BOP. When the upper and lower bounds obtained are equal, the temporary scheduling decision $x^{t*}$ obtained from the Main problem is also optimal to robust BOP, and the algorithm terminates. Otherwise, we update the set $\Omega^t$ to include $\omega^{t*}$ obtained from the Recourse problem,
$\Omega^{t+1}\leftarrow \Omega^t \cup \{\omega^{t*}\},$
and re-solve the updated the Main problem with $\Omega^{t+1}$. 

We formalize the above AC\&CG algorithm for solving robust BOP in Algorithm~\ref{brop-accg} in Appendix~\ref{app:robust_optimization}. Note that each iteration of AC\&CG adds a LOS realization trace, $\omega^{t*}$ to the set $\Omega^t$. This process adds a significant number of new variables and related constraints to the Main problem.\footnote{See Equations \eqref{Main-y1} and \eqref{Main-y2} in Appendix~\ref{app:robust_optimization}.} As the number of iterations increases, the size of the Main problem can therefore grow quickly. When implementing AC\&CG, we thus terminate the algorithm after at most 10 iterations due to computational constraints. In Appendix~\ref{app:additional_numerical_results}, we show that the optimally gaps, $\frac{UB - LB}{LB}$, are reduced to under 1\% for almost all batches after 10 iterations. We discuss the computational challenges further in Section~\ref{Insights}.  

The full rolling robust optimization algorithm (RRO) combining information update and AC\&CG is presented in Algorithm~\ref{rro}.

  \begin{algorithm}
  \caption{Rolling Robust Optimization (RRO)}\label{rro}
  \begin{algorithmic}[1]
\STATE Initialize with $b=1$, $P^{past}_b =\emptyset$, $\tilde{x} = \emptyset$.
\FOR{$b = 1$ \TO $B$}
\vspace{2mm} 
\STATE \textbf{1. Information Update.} Patient arrival 
\STATE $P_b$ is realized between $s_{b-1}$ and $s_b-1$; evaluate $l^{min}_p$, $l^{max}_p$ for all $p\in P_b$.
\FOR{$p \in P^{past}_b$}
\STATE Update LOS distributions and $l^{min}_p$, $l^{max}_p$ using the \textit{Information Update Procedure} for robust BOP.
\ENDFOR

\STATE \textbf{2. Schedule Optimization.} Solve robust BOP with AC\&CG given $s_b$, $P_b$, $P^{past}_b$, $\tilde{x}$, $l^{min}_p, l^{max}_p$ $\forall P\in P_b\cup P^{past}_b$; implement solution $x^*$ for all $p\in P_b$.
\STATE \textbf{3. Schedule and Capacity Update.} 
\STATE $P_{b+1}^{past} \leftarrow P_b^{past} \cup P_b$
\STATE $\tilde{x}\leftarrow \tilde{x}\cup x^*$
\ENDFOR
  \end{algorithmic}
  \end{algorithm}

\section{Numerical Experiments and Results}\label{sec:results}
In this section, we evaluate the three algorithms introduced in Section~\ref{sec:sp} based on numerical simulations with real-world data. Using the dataset and procedures described in Section \ref{sec:los_predict}, we develop a regression model for LOS prediction using GBM on historical training data from 2014 to 2018. We use the machine learning model to generate LOS predictions for all patients in the test set, $\{v_1,v_2,\ldots,v_{|P|}\}$. The empirical distribution of predicative errors on both the training set and the test set\footnote{In practice, prediction errors of the test set will not be available at the time of scheduling. The test errors are included in constructing $\mathcal{S}$ so that, if the LOS of an incoming patient in the test set surpasses all cases in the training set, conditional sampling described in the previous section (see Step 2(b) of Method 1 still works as intended. Although this is not ideal, the scheduling process should not benefit much from it and it helps simplify our simulations. 
In practice, if the LOS of a patient surpasses all previous cases, one approach is to consult health providers.
}, are used as the estimated distribution $\mathcal{S}$. The LOS predictions and $\mathcal{S}$ are then used as inputs to Algorithm~\ref{SRSO}, \ref{CRSO} and \ref{rro} to generate optimal surgical schedules. Performances of the optimal schedules are simulated and evaluated using historical patient arrivals and actual LOS data in the test set from September 2018 to March 2020.

Our numerical results indicate that both Conservative-RSO and RRO outperform the status quo, while Standard-SP fails to achieve consistent performance improvement. Moreover, Conservative-RSO achieves the best overall performance despite its relative computational simplicity compared to RRO. The results highlight the importance of tailoring algorithm for long-tailed distributions and reveals shortcomings of complex algorithms in practical settings. The best-performing algorithm, Conservative-RSO, provides a promising direction for designing efficient CVICU scheduling algorithms.

\subsection{Performance on Historical True LOS}\label{sec:historical-trace}
We first simulate the performance of Standard RSO, Conservative-RSO and RRO on historical patient arrivals. 

Each run of either stochastic optimization algorithms may produce a different scheduling policy, because the objective function involves random sampling. For this reason, both Standard-RSO and Conservative-RSO are run 90 times each on the testing period for each values of $\beta\in\{1,5,10,20\}$, where greater values of $\beta$ mean more weight on the cost of ICU overflow compared to wait time. 

In contrast, the solution of RRO is deterministic for the same set of model parameters. When evaluating RRO, we thus only run the algorithm once for each set of parameters. For RRO, we tune both values of $\beta\in\{1,5,10,20\}$ and the uncertainty budget, $\eta \in \{0.5, 0.75, 1.0\}$. Greater values of $\eta$ mean a larger uncertainty set containing longer LOS realizations. 

Performance of each optimal policy is simulated using the historical true LOS trace, i.e., the actual realized values of LOS for each patient in the testing period. The change in wait times and the number of high ICU occupancy days in the ICU for each experiments are evaluated in comparison to those of the original schedule (i.e. the status quo). Good performance corresponds to an improvement on both metrics.
       \begin{figure*}[!h]
        \centering
        \begin{subfigure}{.5\textwidth}
        \centering
         \includegraphics[width=0.95\linewidth]{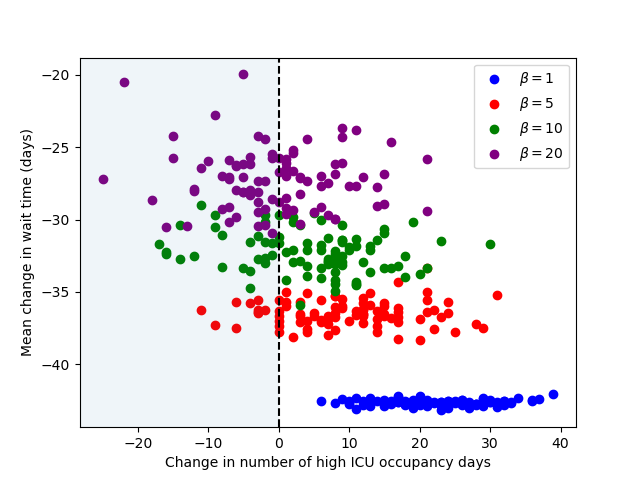}
         \caption{Mean change in wait times vs. ICU congestion}
         \label{fig:fig9a}
        \end{subfigure}%
        \begin{subfigure}{.5\textwidth}
        \centering
         \includegraphics[width=0.95\linewidth]{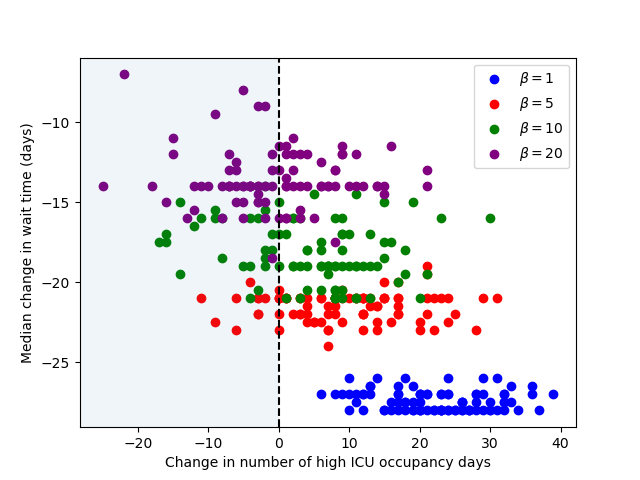}
         \caption{Median change in wait times vs. ICU congestion}
         \label{fig:fig9b}
    \end{subfigure}
    \caption{Performance trade-off of Standard-RSO between patient wait times and ICU congestion using different values of $\beta$, compared to the status quo. ICU congestion is measured using the number of high ICU occupancy days with at least 10 elective patients in the ICU ($u_d \geq 2$).}
    \label{fig_fig9}
    \end{figure*}
  
   \begin{figure*}[!h]
        \centering
        \begin{subfigure}{.5\textwidth}
        \centering
         \includegraphics[width=0.95\linewidth]{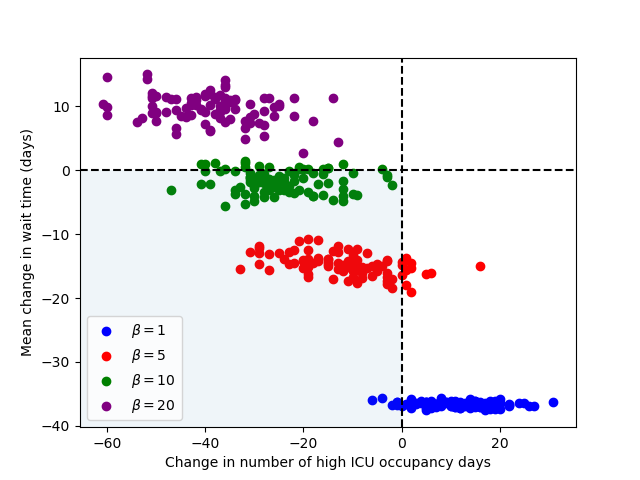}
         \caption{Mean change in wait times vs. ICU congestion}
        \label{fig:fig8a}
        \end{subfigure}%
        \begin{subfigure}{.5\textwidth}
        \centering
         \includegraphics[width=0.95\linewidth]{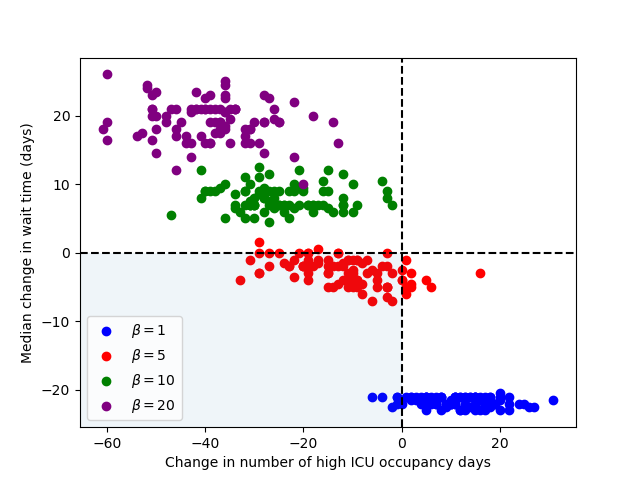}
         \caption{Median change in wait times vs. ICU congestion}
         \label{fig:fig8b}
    \end{subfigure}
    \caption{Performance trade-off of Conservative-RSO between patient wait times and ICU congestion using different values of $\beta$, compared to the status quo. ICU congestion is measured using the number of high ICU occupancy days with at least 10 elective patients in the ICU ($u_d \geq 2$). 
    }
      \label{fig_fig8}
    \end{figure*}

       \begin{figure*}[!h]
        \centering
        \begin{subfigure}{.5\textwidth}
        \centering
         \includegraphics[width=0.95\linewidth]{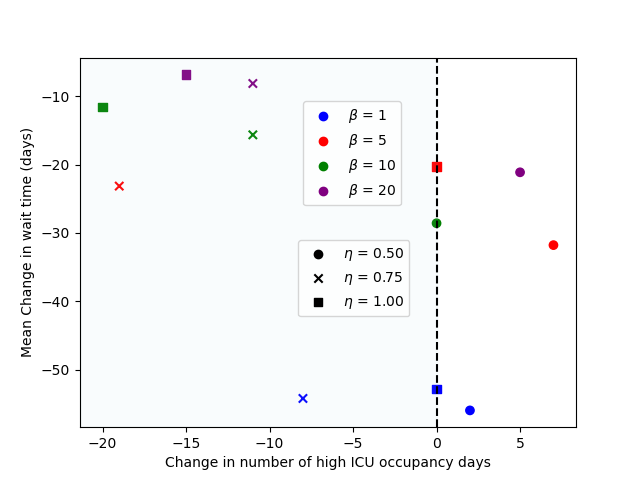}
         \caption{Mean change in wait times vs. ICU congestion}
         \label{fig:fig_robust_a}
        \end{subfigure}%
        \begin{subfigure}{.5\textwidth}
        \centering
         \includegraphics[width=0.95\linewidth]{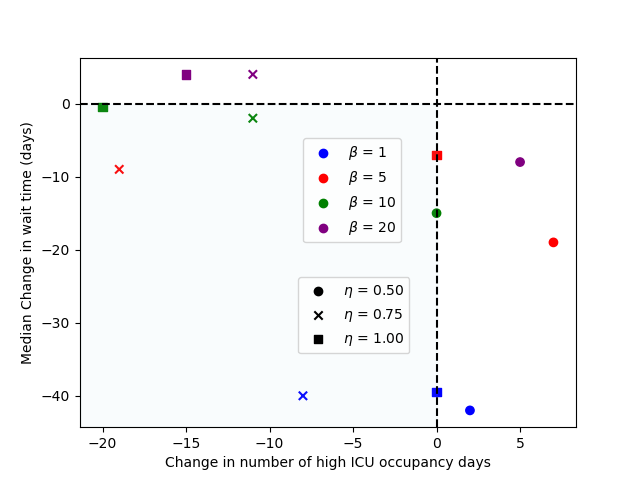}
         \caption{Median change in wait times vs. ICU congestion}
         \label{fig:fig_robust_b}
    \end{subfigure}
    \caption{Performance trade-off of RRO between patient wait times and ICU congestion using different values of $\beta,\eta$ compared to the status quo. ICU congestion is measured using the number of high ICU occupancy days with at least 10 elective patients in the ICU ($u_d \geq 2$).}
    \label{figs_robust}
    \end{figure*}

Performance of the Standard-RSO, Conservative-RSO and RRO are presented in Figure~\ref{fig_fig8}, \ref{fig_fig9} and \ref{figs_robust} respectively. Each run of experiments is plotted using the average or median change in wait time (the y coordinate) and the simulated reduction in high ICU occupancy days in the ICU (the x coordinate). The quadrant shaded in blue in each plot indicates performance improvement on both metrics. Different values of $\beta$ indicated by color, and different values of $\eta$ (for RRO only) are indicated by marker type.

Standard-RSO shows poor performance in reducing ICU overflow regardless of the value of $\beta$,\footnote{For consistency, we present performance for $\beta\in\{1,5,10,20\}$ for all algorithms. It is worth noting that increasing the value of $\beta$ for Standard-RSO does not lead to meaningful improvement in performance. One example is provided in Figure~\ref{figs_sro-bi} of Appendix~\ref{app:additional_numerical_results}.} as illustrated in Figure~\ref{fig:fig9a} and \ref{fig:fig9b}. Despite considering the stochasticity in LOS, Standard-RSO tends to aggressively schedule most patients much earlier and is unable to effectively handle the trade off between ICU overflow and patient wait times. 

In contrast, Conservative-RSO demonstrates consistent improvement of performance in both metrics compared to the status quo in Figure~\ref{fig:fig8a} in terms of the average change in wait time and the reduction in high ICU occupancy days for $\beta\in\{5,10\}$. Relatively weaker performance is observed when the median change in wait times is examined in Figure~\ref{fig:fig8b}. This is likely because the formulation of the objective function specifically minimizes the sum of wait times instead of the median. As expected, higher values of $\beta$ lead to less high ICU occupancy days but longer wait times.

RRO also achieves good performance shown in Figure~\ref{figs_robust}. In particular, performance improvement on both metrics parameter are achieved under parameter combinations $(\eta = 1, \beta=10)$ and $(\eta = 0.75, \beta\in\{5,10,20\})$. Higher values of $\beta$ and uncertainty budget $\eta$ tend to result in less high ICU occupancy days but longer wait times, in line with our expectation.

Although Conservative-RSO and RRO are more conservative in its ICU occupancy forecasts and lead to longer patient wait times, there is enough slack in the original system so that both are able to effectively reduce ICU congestion without excessively pushing back surgeries compared to the original schedule. In addition, the contrast in performance between Standard-RSO and Conservative-RSO further reflects that, with careful design choices targeting the challenges specific to the problem, performance improvement can be achieved in practical settings without increasing complexity of the algorithm

\subsection{Performance on Bootstrapped Traces of LOS}

\begin{figure*}[!h]
        \centering
        \begin{subfigure}{.5\textwidth}
        \centering
         \includegraphics[width=0.95\linewidth]{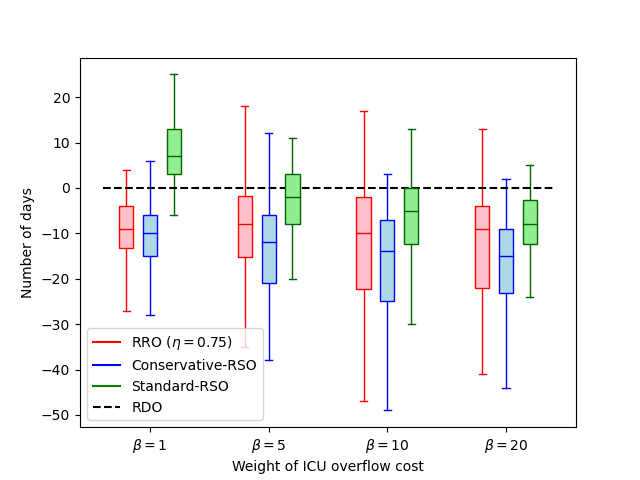}
         \caption{Change in high ICU occupancy days, $O_d \geq 10$}
        \label{fig_fig10a}
        \end{subfigure}%
        \begin{subfigure}{.5\textwidth}
        \centering
         \includegraphics[width=0.95\linewidth]{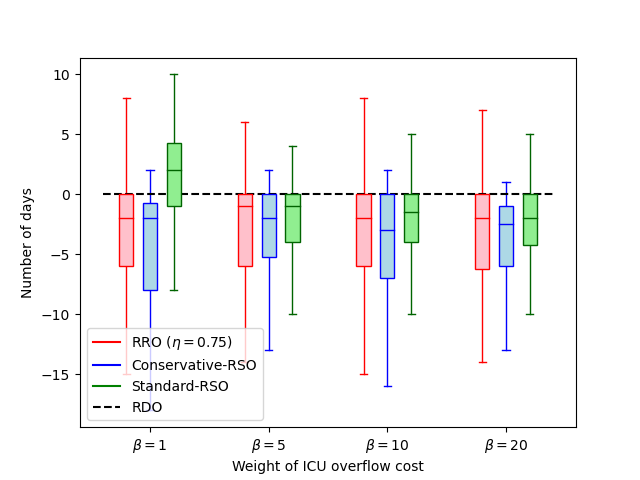}
         \caption{Change in high ICU occupancy days, $O_d \geq 12$}
         \label{fig_fig10b}
    \end{subfigure}
    \caption{Difference in the number of high ICU occupancy days using RRO, Conservative-RSO and Standard-RSO relative to the deterministic benchmark by RDO, for different values of $\beta$.
    High ICU occupancy days in (a) are days with at least 10 elective patients in the ICU,  and those in (b) are days with at least 12 elective patients. Each box plot shows the median, the interquartile range, minimum and maximum of 100 experiments. }
    \label{fig_fig10}
    \end{figure*}

\begin{figure}[!h]
    \centering
    \includegraphics[width=0.95\linewidth]{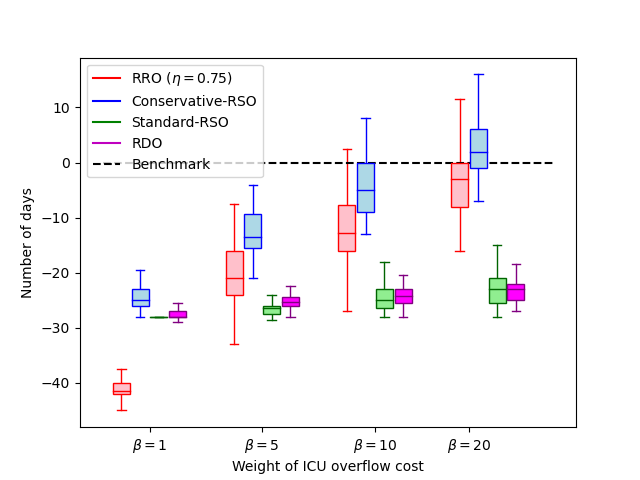}
    \caption{Median change in wait times relative to the status-quo using RRO, Conservative-RSO, Standard-RSO and RDO respectively for different values of $\beta$. }
    \label{fig_fig11}
\end{figure}

In order to obtain confidence intervals for the relative performances of the three methods, we now use bootstrapping \cite{davison1997bootstrap} 
to generate multiple \textit{evaluation traces} of LOS realization for all patient arrivals in the test data. 

Let $\{v_1,v_2,\ldots, v_{|P|}\}$ be the historical sequence of predicted LOS values for all patients to be scheduled during the test period, $P = \bigcup_{b=1}^B P_b$. The procedure of evaluating each method on bootstrapped evaluation traces is as follows.

\paragraph{Performance evaluation with bootstrapping.} Repeat the following steps to sample $N_E$ evaluation traces $\omega'^{(i)}= \{l'^{(i)}_p: p \in P\}$ for $i=1,2,\ldots,N_E$:
\begin{itemize}[label=$\bullet$]
\item Step 1: Generate a bootstrapped patient arrival sequence. The historical arrival sequence is defined by a list of patient arrival times, $\{t_1,t_2,\ldots,t_{|P|}\}$ with predicted LOS $\{v_1,v_2,\ldots,v_{|P|}\}$. We uniformly sample with replacement to obtain a new sequence of arrival with predicted LOS $\{v'^{(i)}_1,v'^{(i)}_2,\ldots, v'^{(i)}_{|P|}\}$ while fixing the arrival times $\{t_1,t_2,\ldots,t_{|P|}\}$.
\item Step 2. Generate predicted LOS distributions for the new arrival sequence using $\{v'^{(i)}_1,v'^{(i)}_2,\ldots, v'^{(i)}_{|P|}\}$ and $\mathcal{S}$. 
\begin{itemize}
    \item Use the new arrival sequence and its predicted LOS distributions as common inputs for Standard and Conservative RSO, and RRO.
    \item Use the same arrival sequence and set $l_p = v'^{(i)}_p \forall p$ as inputs for the deterministic formulation, RDO.
\end{itemize}
Each algorithm produces one optimal schedule $x^*$ for the sequence.
\item Step 3: Generate a trace of realized LOS for the new arrival sequence by randomly sampling from the predicted LOS distributions. Specifically, for patient arrivals at $\{t_1,t_2,\ldots,t_{|P|}\}$, sample the sequence of prediction errors $\{r'^{(i)}_1,r'^{(i)}_2,\ldots,r'^{(i)}_{|P|}\}$ independently from $\mathcal{S}$ and obtain the trace of realized LOS, $$\omega'^{(i)} = \{l'^{(i)}_1,l'^{(i)}_2,\ldots,l'^{(i)}_{|P|}\}$$
by setting  $l'^{(i)}_p = round(v'^{(i)}_p \cdot r'^{(i)}_p)$ for all $p$.
\item Step 4. Simulate the performance of the optimal schedule by each algorithm under the LOS realization scenario described by $\omega'^{(i)}$. Compare relative performance on both metrics. Performance of RDO is used as a benchmark for comparison. 
\end{itemize}

Similar to Section \ref{sec:historical-trace}, Step 1-4 are repeated for 100 iterations ($N_E = 100$)
on each value of $\beta$ for each algorithm. 
For RRO, $\eta=0.75$ is used because of its good performance on the historical trace shown in Figure~\ref{figs_robust}.
We compare the change in wait times relative to the original schedule, and compare the number of high ICU occupancy days of the three algorithms relative to that of deterministic optimization, RDO.

Figure~\ref{fig_fig10} presents the performance of Standard-RSO, Conservative-RSO and RRO in reducing ICU congestion on bootstrapped patient arrivals in comparison to the performance benchmark set by RDO. 

Standard-RSO manages to achieve some overflow reduction for $\beta\geq 10$. When $\beta$ is small, however, Standard-RSO tends to under-perform compared to the deterministic benchmark. In contrast, the interquartile ranges of Conservative-RSO (blue) and RRO (red) both show more significant reduction in the number of high ICU occupancy days compared to RDO for all values of chosen $\beta$.

The \textit{median} of changes in patient wait time compared to the status quo are presented in Figure~\ref{fig_fig11}. Standard-RSO demonstrates similar behaviors to its deterministic counterpart, with the tendency to aggressively schedule patients earlier. In contrast, Conservative-RSO strategically delays procedures so that ICU overflow can be effectively avoided. For $\beta \in \{1,5,10\}$, the Conservative-RSO is able to consistently reduce ICU overflow without pushing back a majority of procedures compared to the original calendar. RRO, on the other hand, achieves comparable performance to Conservative-RSO and is able to reduce ICU overflow without pushing back procedures for all values of $\beta$.

Comparing Conservative-RSO and RRO, both algorithms utilize conservative LOS estimates to address under-predictions for long-tailed LOS. Both achieves promising reductions in ICU congestion without excessive increase in wait times. However, Conservative-RSO demonstrates better average and worst-case performance (upper caps of the box plots) in ICU overflow for all selected values of $\beta$.

The relative poorer performance of RRO in reducing ICU overflow may be a result of its sensitivity towards the choice of other parameters, such as the uncertainty budget $\eta$, the choice of $l^{min}_p$ and the percentile $1-\alpha$ used for setting $l^{max}_p$. During our effort to calibrate the uncertainty set, the performance of RRO is observed to be drastically different under different parameter choices. While there remains room for potential performance improvement with better parameter tuning, finding the optimal combination of all parameters can be a very difficult task in practice given the large number of possibilities and the long computation time required for each run of the AC\&CG algorithm. This reveals a shortcoming of using robust optimization approaches that involve a large number of parameters, and demonstrates that increasing algorithmic complexity does not always leads to better performance in practice. To improve practicality of robust optimization, further research is needed to develop new formulations less sensitive to parameter choices. 

Lastly, we note that there is still a notable performance gap between our best-performing model, \\ Conservative-RSO, and the theoretical upper bound established in Section \ref{sec:determ} with rolling deterministic optimization under perfect information. This implies significant room for further development of more efficient data-driven optimization methods, with Conservative-RSO serving as a stepping tone towards a promising direction.

\subsection{Summary of numerical results and additional approaches}
We examined a variety of optimization formulations in conjunction with data-driven LOS predictions. All approaches are developed under similar frameworks combining machine learning, rolling information update and optimization methods, but they differ significantly in their simulation performances, strengths and weaknesses. We provide a summary of these algorithms in Table~\ref{table:summary}.
\begin{table*}[!h]
\centering
\begin{tabular}{|c|c|c|}
\hline
\textbf{Algorithm}  & \textbf{Performance}  & \textbf{Challenges}                         \\ \hline
RDO                     & \begin{tabular}[c]{@{}c@{}}Does not reduce ICU congestion \\ without increasing patient wait times\end{tabular}       & \begin{tabular}[c]{@{}c@{}}Weak predictive accuracy for\\ prolonged LOS\end{tabular}    \\ \hline
Standard-RSO                     & \begin{tabular}[c]{@{}c@{}}Does not reduce ICU congestion \\ without increasing patient wait times\end{tabular}       & \begin{tabular}[c]{@{}c@{}}Insufficient sampling on rare \\ occurrences of prolonged LOS\end{tabular}                            \\ \hline
Conservative-RSO           & \begin{tabular}[c]{@{}c@{}}Promising reductions in both \\ ICU congestion and wait time\end{tabular}           & \begin{tabular}[c]{@{}c@{}}Can be overly conservative \\ without precise LOS prediction\end{tabular}     \\
\hline
RRO     & \begin{tabular}[c]{@{}c@{}}Promising reductions in both \\ ICU congestion and wait time\end{tabular}         & Slow and difficult to tune                           \\ 

\hline
\end{tabular}
\caption{Comparison of different algorithms}
\label{table:summary}
\end{table*}

Numerous alternative formulations of each of the these approaches were examined:
\begin{itemize}[label=$\bullet$]
\item For stochastic optimization, estimating the distribution of true LOS $L_p$ using the historical empirical distribution of LOS within the same procedure type, instead of sampling prediction errors.
\item For robust optimization, using an uncertainty budget that scales with $\sqrt{|A_b|}$ instead of $|A_b|$ (see Equation~\eqref{recourse-l2}), and various values of $1-\alpha$ for the LOS upper bound, including $\alpha\in\{0.1,0.15,0.2,0.25,0.3\}$.

\item Various alternative optimization formulations, such as linear $f(u_d)$, penalizing under-utilization of OR blocks or minimizing maximum wait times, etc.

\item Various alternative machine learning models. In particular, We used H2O AutoML \cite{H2OAutoML20} to search for other promising models from its wide range of built-in machine learning and deep learning algorithms.
\end{itemize}

None of the above alternatives result in consistent performance improvements compared to results presented in this section. 

\section{Insights and Discussions}\label{Insights}

Among all the practical formulations considered,\\Conservative-RSO and RRO both
achieved reduction in both ICU congestion and patient wait times. The key driver of their promising performance is the focus on under-estimates of long-tailed, extended LOS of cardiovascular surgery patients, through design of either SAA sampling of LOS realizations, or the uncertainty set of LOS. Good performance also relies on precise choices of the trade-off coefficient $\beta$, and several other model parameters in the case of robust optimization. To achieve the desired optimization performance in practice, all model parameters need to be carefully tuned to balance reduced ICU congestion with longer wait times. 

Moreover, on each scheduling day, possible LOS outcomes for patients currently in the ICU are updated given how long they have stayed in the ICU. In particular, if some patients are already staying longer than expected, their projected probability distribution of LOS will be updated accordingly based on the new information. This information update procedure further corrects for potential under-estimates of prolonged LOS at the time of scheduling and improves optimization performance.

The fact that the optimization outcomes are highly sensitive to these nuanced design choices underscores the operational challenges of surgical scheduling with long-tailed LOS. Any solution will need to be tailored and fine-tuned to suit the context of specific institutions in order to reach the desired outcome.

Although our proposed scheduling framework has demonstrate promising performance, there remains room for potential extensions and future work to address some of its limitations.

First, under our rolling optimization framework, patients who arrived in-between two scheduling days need to wait until the next scheduling day to be scheduled. In our simulations, we have chosen the scheduling days $s_b$ to be one month apart for elective procedures based on our context, as explained at the end of section~\ref{sec:batch}. However, for other types of procedures, scheduling may need to happen on a daily bases. While our modeling framework naturally extends to the case of daily scheduling (by setting $s_b$ as consecutive days), managing ICU overflow in this case can become more difficult, since scheduling is performed with less patient arrival information. This is a limitation of our framework to be addressed in future works. For instance, authors of \cite{keyvanshokooh2022coordinated} propose an alternative rolling horizon framework for surgical scheduling that takes into account uncertain future patient arrivals. With a forward-looking, rolling arrival horizon, scheduling decisions can be made more frequently for realized patient arrivals, in anticipation of stochastic future arrivals. Incorporating designs of this kind with our current framework has the potential to improve practicality and performance for procedures that require more urgent scheduling.

Second, a challenge of both Conservative-RSO and RRO that has not been addressed is their computational limitations. In our numerical experiments, we have limited the number of sampled traces, $N_\Omega$ for SAA to $10$ and the number of iterations of AC\&CG to $11$. As mentioned in Section~\ref{sec:rsp} and \ref{sec:rro}, the memory and run time required to solve the optimal schedule for each batch of patients can grow quickly when $N_\Omega$ and the number of AC\&CG iterations increase. In the case of SAA for Standard and Conservative-RSO, it typically took 4 to 8 hours to run each numerical experiment for $N_\Omega = 10$ with 19 batches with two CPUs and 16GB memory per CPU. For $N_\Omega=30$, at least 21 hours were required to run one such numerical experiment with four CPUs and 16GB of memory per CPU. Meanwhile, AC\&CG for RRO requires even longer run time: it took typically 7 to 12 hours to run each experiment for 11 iterations with two CPUs and 16GB memory per CPU. The authors of \cite{robust_opt} also point out that the proposed algorithm may be computationally inefficient for larger uncertainty sets. Developing a better understanding of how the choice of $N_\Omega$ and the maximum number of AC\&CG iterations impacts optimization performance is a meaningful subject of further research.

\section{Concluding Remarks} \label{Conclusion}
Using seven years of data from an academic medical center, we developed machine learning models to predict post-surgical length of stay, optimization models to schedule procedures to minimize post-surgical bed congestion, and simulated the results of the use of these models on the most recent 18 months of held-out empirical test data. We established an upper bound on performance with an offline optimization model of historical data with access to actual LOS. 

A conservative stochastic program with sufficient sampling of the LOS distribution to capture tail behavior managed to achieve the best overall performance in reducing ICU congestion without increasing wait times for surgery.
Compared to the hospital's current paper-based system, the deterministic and standard stochastic optimization approaches, along with numerous variants of machine learning and optimization formulations, failed to improve performance. The failure of most models to improve over the status quo, illustrate the importance of using empirical data, rather than synthetic parametric data, especially for long-tailed distributions. Several lessons for dealing with such data are offered by the contrast between the models that did and did not improve performance.

The prevalent null results highlight the challenge of the unpredictability and non-stationarity of prolonged LOS in practical settings. This challenge is rarely addressed in previous works, most of which follow the common practice of using synthetic LOS data generated based on strong distributional assumptions. For example, in \cite{dro} and \cite{min_yih}, the distribution of the LOS in the ICU is assumed to have a mean and standard deviation of no more than 3.5 days. In the robust optimization approach discussed in  \cite{robust_opt}, the worst-case deviation of the uncertainty set is assumed to be 4 days, which the authors admit can be far from reality. Despite the positive results obtained in simulation, the actual performance of these optimization models can be drastically different in practice in the presence of significant LOS estimation errors. 

We proposed a novel formulation of stochastic optimization, Conservative-RSO, which demonstrated most consistent improvements with empirical data despite of its relative algorithmic simplicity. Our results show that Conservative-RSO is a promising first step towards addressing the unpredictability and long-tailed behavior of LOS to improve surgical scheduling, and they provide meaningful directions for future work. First, much work is needed to explore and evaluate alternative designs of data-driven optimization formulations, calibrated and evaluated using empirical data of LOS. When empirical data are not available, any distributional assumptions should incorporate the long-tailed, non-stationary nature of LOS distributions to minimize the bias present in using standard distributional assumptions. Furthermore, learning over long-tailed distributions has not been fully examined in prior research. We highlight its central importance in tackling the general problem of surgical scheduling with constrained downstream capacities. Obtaining accurate and precise LOS predictions is one way to push the Pareto frontier of optimization closer to the theoretical performance upper bound using full information. Settings where patient characteristics available at the time of scheduling fail to explain the majority of variation in LOS offer exciting challenges to be tackled by innovative approaches that combine the power of prediction, optimization and information update.

\bibliographystyle{spbasic}   

\bibliography{bibliography}   

\newpage
\textcolor{white}{test}

\newpage
\appendix
\section{Mathematical Formulations for $f(u_d),Q^{op}$}
\label{app:math_formulation}

Solving the quadratic objective, $f(u_d) = u_d^2$ of the mixed-integer program can be slow. To reduce the runtime required to solve the problem, we implement the quadratic term in the objective function, $f(u_d) = u_d$, using a piece-wise linear approximation,
\begin{equation}\label{eqn:f-piece}
    f(u_d)=e_1 u^{(1)}_d +e_2  u^{(2)}_d + e_3 u^{(3)}_d + e_4 u^{(4)}_d + e_5 u^{(5)}_d,
\end{equation}
and add additional constraints 
$$\sum_{p\in  P_b \cup P^{past}_b} y_{d,p}\leq c + m - 1 +  u^{(m)}_d \quad \forall d,m$$
$$u^{(m)}_d \geq 0\quad \forall d, m$$

In words, $u_d^{(m)}$ counts ICU overflow above $c+m-1$ for $m=1,\ldots,5$. Here, $e_m$ are constant coefficients for the piecewise linear function. 

In our formulation, we set $e_1 = 1, e_2=e_3=e_4=e_5=2$. Since $u_d$ only takes integer values, this coefficient choice leads to $f(u_d)\equiv u_d^2$ for any $u_d\leq 5$. 
The piece-wise linear approximation is 20 times faster than the quadratic form in our numerical experiments. 

The above approximation is used in all our algorithms. The formulations for stochastic and robust algorithms are analogous (e.g. copies of $u^(m)_d$ are created for different LOS realizations), and we thus omit the details here. 

Next, we introduce the full mathematical formulation of $Q^{op}$ in constraint \eqref{con:operations} for offline optimization below.\\

\textbf{Sets}

 \begin{list}{$\bullet$}{}  
 \item $D = \{1,2,\ldots,N_d\}$: index for days
 \item $P = \{1,2,\ldots,N_p\}$: index for patients
 \item $P^{par} \subset P$: set of PAR patients
 \item $K$: set of surgeons
 \end{list}

\textbf{Parameters}

 \begin{list}{$\bullet$}{} 
 \item $c$: number of ICU beds reserved for elective patients, set to 8
 \item $q_p$: operation duration for $p \in P$. 
 \item $l_p$: post-op length of stay in CVICU for $p \in P$
 \item $h_{d,k}$: number of hours surgeon $k$ is available to perform surgery on day $d$, for $s \in S$. $h_{d,k} = 15$ if $k$ operates on day $d$, 0 otherwise.
 \item $par\_day_{d,k}$: indicator variable for PAR days. $par\_day_{d,k} = 1$ 
 if surgeon $k$ can perform PAR surgeries on day $d$, 0 otherwise.
  \item $original\_date_{p} \in D$: the original date that a surgery is scheduled for for $p \in P$
\item $actual\_date_{p} \in D$: the actual date that a surgery takes place for $p \in P$; can be different from $original\_date_{p}$ if the surgery was rescheduled.
\item $arrived\_on_{p}$: the arrival date for $p \in P$, i.e., when a patient is first being scheduled for surgery; can be negative (i.e. outside the set $D$) if the surgery was scheduled before September 2018

 \item $lead_{p}$: the lead time of a surgery, i.e.  
 \begin{align*}
 original\_date_{p}- arrived\_on_{p}
 \end{align*}

  \item $m_{k,p}$: binary, 1 if patient $p$ is assigned to surgeon $k$ and 0 otherwise
 \end{list}

\textbf{Decision Variables}
 
 \begin{list}{$\bullet$}{}
 \item $u_d$: integer, the number of additional elective patients in the ICU on day $d$ above $c$
\item $ x_{d,p}$: binary, $ x_{d,p} = 1$ if patient $p$ is scheduled to have her surgery on day $d$; otherwise 0
\item $d_p$: the date when the surgery of patient $p$ is scheduled by the model
\item $y_{d,p}$: binary, $y_{d,p} = 1$ if patient $p$ stays in the CVICU on day $d$; otherwise 0

\item $z_{d,p}$: continuous, number of hours that the surgery of $p$ lasts on day $d$. $z_{d,p} = q_p$ if $x_{d,p} = 1$, otherwise 0

\end{list}

\textbf{Indicator Function}

 \begin{list}{$\bullet$}{}
 \item Assumed feasible window where patient $p$ is available for surgery:
\begin{equation*}
    g(d,p) = \mathbbm{1}\{d^{min}_p \leq d<d^{max}_p\}
\end{equation*}
where $d^{max}_p = \text{min}(actual\_date_p+90,N_d)$,
\begin{equation}
d^{min}_p =
    \begin{cases}
    \text{max}[1,original\_date_p- \frac{lead_p}{2}], & lead_p> 20\\
    \text{max}[1,arrived\_on_{p}],&lead_p\leq20
    \end{cases}
\end{equation} 
\label{sec:offline_optimization}
\end{list}

Next we describe the optimization constraints (\ref{constr2})-(\ref{constr6}) that are equivalent to $x\in Q^{op}.$

Constraints (\ref{constr2}) ensures each patient is scheduled exactly once within that patient's window of availability. 
\begin{equation}
    \sum_{d\in D} x_{d,p}\cdot g(d,p) = 1,\quad\forall p\in P
        \label{constr2}
\end{equation}
Constraints (\ref{constr3}),(\ref{constr4}),(\ref{constr5}) (\ref{constr6}) incorporate daily availability of surgeons. Constraint (\ref{constr3}) includes number of hours a surgeon is available on each day for (non PAR) surgeries.
\begin{equation}
z_{d,p} = x_{d,p}q_p, \quad \forall d\in D, p\in P
\label{constr3}
\end{equation}
\begin{equation}
\sum_{p\in P}z_{d,p} m_{k,p} \leq h_{d,k}, \quad \forall d\in D, k\in K
\label{constr4}
\end{equation}
Constraint (\ref{constr5}) captures that PAR surgeries can only be done on pre-specified days.
\begin{equation}
\sum_{p\in P^{par}}x_{d,p} m_{k,p} \leq par\_day_{d,k}, \quad \forall d\in D, k\in K
\label{constr5}
\end{equation}
Constraint (\ref{constr6}) ensures each surgeon is not scheduled for PAR surgeries in subsequent days.
\begin{equation}
\begin{aligned}
\sum_{p \in P^{par}}x_{d,p}m_{k,p} + \sum_{p \in P^{par}} & x_{d+1,p}m_{k,p}  \leq 1,\\  &\forall d\in D\setminus\{N_d\},k\in K
\end{aligned}
\label{constr6}
\end{equation}

\subsection{$Q^{op}$ for Batch Optimization Problem}
\label{app:batch_deterministic_formulation}
When solving BOP for rolling scheduling, $Q^{op}$ is adjusted accordingly. In addition to the Sets and Parameters specified in Section \ref{app:math_formulation}, we use the following sets and parameters.

\textbf{Sets}

 \begin{list}{$\bullet$}{}  
 \item $P_b \subset P$: index of batch patients of period $b$
 \item $P^{past}_b = \bigcup_{k=1}^{b-1} P_k$: set of patients prior to period $b$
 \item $P_b^{par} = P^{par} \cap P_b$
 \end{list}

\textbf{Parameters}

 \begin{list}{$\bullet$}{} 
 \item   $s_b$: start date/day of batch $b$ 
 \item   $d^{min}_{p,b} = \max(d^{min}_p,s_b)$
 \item $x^*_{d,p}$: solutions obtained from previous periods
 \end{list}
 
Since the definition of $d^{min}_{p,b}$ may restrict the original time window of availability for some patients, we also adjust the last available date, $d^{max}_p$, to at least 90 days after the start of the period, i.e.,
\begin{equation}\label{const:latest_date_1m}
  d^{max}_{p,b} =  \max(d^{max}_p,\min(s_b+90,N_d))
\end{equation}
This adjustment allows 
flexible scheduling as described in Section~\ref{section:offline}. In practice, the definition of $d^{max}_{p,b}$ will not include $N_d$; we included it for our simulation runs. Although the latter adjustment could potentially increase wait time, it is necessary in ensuring that the set of feasible scheduling solutions is not too restricted, and any resultant increase in wait time will be penalized by the objective function.  

\textbf{Indicator Function}
 \begin{list}{$\bullet$}{}
 \item Adjusted feasibility window for each patient.
 \end{list}
\begin{equation*}
    g_b(d,p) = \mathbbm{1}\{d^{min}_{p,b} \leq d< d^{max}_{p,b}\}
\end{equation*}

Constraints \eqref{eqn:bop-op1}-\eqref{eqn:bop-op2} give the equivalent formulation of $x\in Q^{op}$ in all deterministic, stochastic and robust BOPs.

\begin{equation}\label{eqn:bop-op1}
    \sum_{d\in D} x_{d,p}g_b(d,p) = 1,\quad\forall p\in P_b
\end{equation}
\begin{equation}
z_{d,p} = x_{d,p}q_p, \quad \forall d\in D, p\in P_b
\end{equation}
\begin{equation}
\sum_{p\in P_b}z_{d,p} m_{s,p} \leq h_{d,s}, \quad \forall d\in D, s\in S
\end{equation}
\begin{equation}
\sum_{p\in P^{par}_b}x_{d,p} m_{s,p} \leq par\_day_{d,s}, \quad \forall d\in D, s\in S
\end{equation}
\begin{equation}\label{eqn:bop-op2}
\begin{aligned}
\sum_{p \in P^{par}_b}x_{d,p}m_{s,p} + \sum_{p \in P^{par}_b} & x_{d+1,p}m_{s,p}  \leq 1,\\  &\forall d\in D\setminus\{N_d\},s\in S.
\end{aligned}
\end{equation}

\section{Solving AC\&CG For Robust BOP}\label{app:robust_optimization}

In the following, we introduce the AC\&CG algorithm used for solving the robust BOP.

\paragraph{1. Main problem.} We start with a subset of traces in the uncertainty set, $\Omega^t = \{\omega^{(n)}:n= 1,\ldots,|\Omega^t|\}\subseteq \mathcal{U}_b$. Instead of minimizing the worst-case cost over the entire uncertainty set $\mathcal{U}$, the Main problem of AC\&CG minimizes the worst-case cost over the subset, $\Omega^t$:
\begin{subequations}\label{eqns:accg-Main}
\begin{equation}\label{obj:Main}
    \min_{x}\quad \sum_{p\in P_b}
    \sum_{d=s_b}^{N_d}(d - d^{min}_p)^+ x_{d,p} + \beta \cdot \theta
\end{equation}
 s.t.
\begin{equation}
x_{d,p} = \tilde{x}_{d,p}\quad \forall p\in P^{past}_b, d\in D
\end{equation}
\begin{equation}
    \sum_{d=s_b}^{N_d} x_{d,p} = 1,\quad \sum_{d=1}^{s_b -1} x_{d,p}=0 \quad \forall p\in P_b 
\end{equation}
\begin{equation}\label{Main-y1}
\begin{aligned}
y^{(n)} = \sum_{d'=\max(d-l^{(n)}_p+1,1)} x_{d',p} &\quad \forall p \in P_b\cup P^{past}_b \\ & \forall d\in D, n=1, \ldots,|\Omega^t|
\end{aligned}
\end{equation}
\begin{equation}\label{Main-y2}
    \sum_{p\in P_b \cup P^{past}_b} y^{(n)}_{d,p} \leq c + u^{(n)}_d \quad \forall d\in D, n=1,\ldots,|\Omega^t| 
\end{equation}
\begin{equation}
x\in Q^{op}, x_{d,p}\in\{0,1\}
\end{equation}
\begin{equation}\label{con:Main-theta}
    \theta\geq \sum_{d=s_b}^{N_d}f(u^{(n)}_d) \quad \forall n= 1,\ldots,|\Omega^t|.
\end{equation}
 \end{subequations}
 
Constraint~\eqref{con:Main-theta} sets the variable $\theta$ to be equal to the maximum cost of overflow among all traces of LOS realization, $\omega^{(n)}$. The objective~\eqref{obj:Main} thus minimizes the weighted sum of total wait time and the maximum cost of overflow for all $\omega^{(n)}\in \Omega^t$. The remaining constraints mirror those for BSOP-SAA in Equations~\eqref{eqns:saa}, only replacing the set of sampled traces with set $\Omega^t$. The optimal objective value of the Main problem provides a \textit{lower bound} to the optimal objective value of robust BOP in \eqref{eqns:brop}.

\paragraph{2. Recourse problem.} At each iteration $t$, after the Main problem is solved with $\Omega^t$, an optimal solution $x^{t*}$ is obtained. The Recourse problem aims to find a trace $\omega^{t*}\in \mathcal{U}_b$ that maximizes the cost of ICU overflow under the given schedule $x^{t*}$. Since $x^{t*}$ and $\omega^{t*}$ are feasible under the original BOP, solving the Recourse problem finds an \textit{upper bound} of the objective value of robust BOP.
\begin{subequations}\label{eqns:accg-recourse}
\begin{equation}
\max_{\omega = \{l_p: p\in P_b\cup P^{past}_b\}} \quad \min_{y,u} \sum^{N_d}_{d=s_b} f(u_d)
\end{equation}
s.t.
\begin{equation}\label{recourse:y}
   \begin{aligned}
   y_{d,p} = \sum_{d'=\max(d-l_p+1,1)}^d x^{t*}_{d',p} \,\, \forall p\in P_b\cup P^{past}_b, d\in D \\ 
   \end{aligned} 
\end{equation}
   \begin{equation}
    \sum_{p\in P_b \cup P^{past}_b} y_{d,p} \leq c + u_d \quad \forall d\in D
\end{equation} 
\begin{equation}
y _{d,p}\in\{0,1\}, u_d\geq 0.
\end{equation}
\begin{equation}\label{recourse-l1}
    l_p^{min} \leq l_p\leq l_p^{max}\quad \forall p \in P_b \cup P^{past}_b
\end{equation}
\begin{equation}\label{recourse-l2}
  \sum_{p\in A_b}\left[ \frac{l_p - l_p^{min}}{l^{max}_p - l_p^{min}}\right]\leq \eta\cdot |A_b|.
\end{equation}
\end{subequations}
This formulation follows from Equations~\eqref{brop-theta-start}-\eqref{brop-theta-end}, and constraints~\eqref{recourse-l1} and\eqref{recourse-l2} use the definition of $\mathcal{U}_b$ in Equations \eqref{con:u1} and \eqref{con:u2}.

One key difficulty of solving the Recourse problem in its current form is that the decision variable $l_p$ appears in the boundary of the summation in constraint \eqref{recourse:y}. We follow the reformulation approach in \cite{robust_opt} and extend it to convex, piece-wise linear forms of $f(u_d)$. The reformulated Recourse problem is provided in Appendix~\ref{app:inner_optimization}. We present the full AC\&CG algorithm in Algorithm~\ref{brop-accg}.

  \begin{algorithm}
  \caption{AC\&CG for Robust BOP}\label{brop-accg}
  \begin{algorithmic}[1]
\STATE Initialize with $x^* = \mathbf{0}$, $t=1$, $LB = - \infty$, $UB = \infty$, $\Omega^1 = \{\omega^{(1)}\}$, where $\omega^{(1)} = \{l^{min}_p: p\in P_b\cup P^{past}_b\}$.
\vspace{1mm}
\WHILE{$t \leq T$ \AND $UB-LB > \epsilon$}
\vspace{1mm} 
\STATE
\textbf{1. Main Problem.} 
\STATE Solve the Main problem in \eqref{eqns:accg-Main} with $\Omega^t$; Obtain optimal solution $x^{t*}$ with objective value $\mu^{t*}$.
\vspace{1mm} 
\STATE Update $x^*\leftarrow x^{t*}$, $LB \leftarrow \mu^{t*}$.
\vspace{1mm} 
\STATE \textbf{2. Recourse Problem.} 
\STATE Solve the Recourse problem in \eqref{eqns:accg-recourse} with $x^{*}$; Obtain optimal solution $\omega^{t*}$ with objective value $\nu^{t*}$. 
\vspace{1mm} 
\STATE Update $UB \leftarrow \min \{UB, \nu^{t*}\}$.
\vspace{2mm} 
\IF{$UB-LB > \epsilon$}
\STATE $\Omega^{t+1} \leftarrow \Omega^{t}\cup\{\omega^{t*}\}$
\STATE $t \leftarrow t+1$
\ENDIF
\ENDWHILE
\vspace{1mm} 
\RETURN{$x^{*}$}
  \end{algorithmic}
  \end{algorithm}

\subsection{Solving the Recourse Problem}\label{app:inner_optimization}
The Recourse problem, denoted as $Q(x^{t*})$, involves constraints including $l_p$ decision variables in the boundary of summations. Here, we introduce our MIP reformulation that is readily solvable by Gurobi. We refer readers to \cite{robust_opt} for more details and proofs of its validity. 

As in \cite{robust_opt} we define the variables $v_{d,p},w_{d,p} \in \{ 0,1\}$ for $d \in D, p \in P$. The variable  $v_{d,p}$ is $1$ only if patient $p$ is admitted in the ICU by $d$. Given the temporary solution $x^{t*}$, $v_{d,p}$ are constant parameters determined by  $x^{t*}$. The decision variable $w_{d,p}$ is $1$ only if patient $p$ leaves the ICU by day $d$. So, $y_{d,p} = v_{d,p} - w_{d,p}$.

The inner minimization problem of $Q(x^{t*})$ is
$$\min_{u\geq 0} \sum_{d\in D} e_1 u^{(1)}_d +e_2  u^{(2)}_d + e_3 u^{(3)}_d + e_4 u^{(4)}_d + e_5 u^{(5)}_d$$
$$\sum_{p\in  P_b \cup P^{past}_b} (v_{d,p} - w_{d,p}) \leq c + m - 1 +  u^{(m)}_d \quad \forall d,m
$$
where $e_1 = 1, e_2=e_3=e_4=e_5 = 2$.

We apply strong duality to reformulate the inner maximization as a maximization problem and also substitute for the definition of uncertainty set $\mathcal{U}_b$. Let $d_p$ denote the date of scheduled procedure for patient $p$ according to $x^{t*}$, i.e., $d_p = \sum_{d\in D} d \cdot x^{t*}_{d,p}$. $Q(x^{t*})$ is reformulated below with decision variables $\lambda_d^{(m)}$ and $w_{d,p}$.
\begin{subequations}
\begin{equation*}
\max_{\lambda,w} \sum_{d\in D}\sum_{m=1}^5\left[\sum_{p\in P_b \cup P^{past}_b}(v_{d,p} - w_{d,p})-c-m+1)\right] \lambda_d^{(m)}
\end{equation*}
\begin{equation*}
d_p = \sum_{d\in D} d \cdot x^{t*}_{d,p}\quad \forall p \in P_b \cup P^{past}_b
\end{equation*}
\begin{equation*}
w_{d,p}\geq 1, \quad p\in P_b \cup P^{past}_b, d = d_p + l^{max}_p,\ldots,T
\end{equation*}
\begin{equation*}
w_{d,p}\leq 0, \quad  p\in P_b \cup P^{past}_b, d = 0,\ldots,d_p + l^{min}_p - 1
\end{equation*}
\begin{equation*}
w_{d,p} \leq w_{d+1,p}\quad \forall d\in D, p\in P_b \cup P^{past}_b
\end{equation*}
\begin{equation*}
\sum_{d\in D} (v_{d,p} - w_{d,p}) = l_p, \quad \forall p\in P_b\cup P_b^{past} \,\,\text{where}\,\, p \notin A_b
\end{equation*}
\begin{equation*}
  \sum_{p\in A_b}\left[\frac{\sum_{d\in D} (v_{d,p} - w_{d,p}) - l_p^{min}}{l^{max}_p - l_p^{min}}\right]\leq \eta\cdot |A_b|.
\end{equation*}
\begin{equation*}
0\leq\lambda_d^{(m)}\leq e_m \quad \forall d, m
\end{equation*}
\begin{equation*}
w_{d,p}\in\{0,1\},\quad \forall d,p
\end{equation*}
\end{subequations}
For the optimal solution, we must have $\lambda^{(m)}_d\in\{0,e_m\}$.

The formulation above involves a bilinear term, $w_{d,p}\lambda^{(m)}_d$. Since $w_{d,p} \in \{0,1\}$, we can reformulate the problem by using $q^{(m)}_{d,p} = w_{d,p}\lambda^{(m)}_d$ for all $d,p,m$. 
The final reformulation of the Recourse problem is the following.
\begin{align*}
  Q(x^{t*}) = & \max_{\lambda,w}  \sum_{d\in D}\sum_{m=1}^5\sum_{p\in P_b \cup P^{past}_b} v_{d,p}\lambda^{(m)}_d  -  \nonumber \\ 
    & \sum_{d\in D}\sum_{m=1}^5\sum_{p\in P_b \cup P^{past}_b} q^{(m)}_{d,p} - \nonumber \\
    &  \sum_{d\in D}\sum_{m=1}^5 (c+m-1)\lambda_d^{(m)}
\end{align*}
\begin{equation*}
d_p = \sum_{d\in D} d \cdot x^{t*}_{d,p}\quad \forall p \in P_b \cup P^{past}_b
\end{equation*}
\begin{equation*}
    w_{d,p}\geq 1, \quad p\in P_b \cup P^{past}_b, d = d_p + l^{max}_p,\ldots,T
\end{equation*}
\begin{equation*}
    w_{d,p}\leq 0, \quad  p\in P_b \cup P^{past}_b, d = 0,\ldots,d_p + l^{min}_p - 1
\end{equation*}
\begin{equation*}
w_{d,p} \leq w_{d+1,p}\quad \forall d\in D, p\in P_b \cup P^{past}_b
\end{equation*}
\begin{equation*}
\sum_{d\in D} (v_{d,p} - w_{d,p}) = l_p, \quad \forall p\in P_b\cup P_b^{past} \,\,\text{where}\,\, p \notin A_b
\end{equation*}
\begin{equation*}
  \sum_{p\in A_b}\left[\frac{\sum_{d\in D} (v_{d,p} - w_{d,p}) - l_p^{min}}{l^{max}_p - l_p^{min}}\right]\leq \eta\cdot |A_b|.
\end{equation*}
\begin{equation*}
q^{(m)}_{d,p} \geq \lambda^{(m)}_t-e_m(1-w_{d,p}) \quad \forall d,p,m
\end{equation*}
\begin{equation*}
q^{(m)}_{d,p} \leq e_m w_{d,p}, \,\, q^{(m)}_{d,p} \leq \lambda^{(m)}_d \quad \forall d,p,m
\end{equation*}
\begin{equation*}
q^{(m)}_{d,p} \geq 0, \lambda_d^{(m)} \in \{0, e_m\}, w_{d,p}\in\{0,1\} \quad \forall d,p,m.
\end{equation*}

\section{Additional Numerical Results}\label{app:additional_numerical_results}
This section presents additional results on the optimality gap of AC\&CG and biweekly scheduling.
\begin{figure}[!h]
    \centering
    \includegraphics[width=1\linewidth]{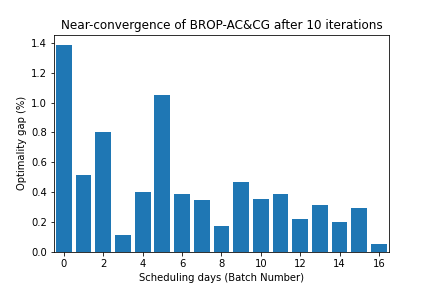}
    \caption{The optimality gap, $\frac{UB-LB}{LB}$, of most batches are below 1\% after 10 iterations.}
    \label{fig:accg-gap}
\end{figure}

       \begin{figure*}[!h]
        \centering
        \begin{subfigure}{.5\textwidth}
        \centering
         \includegraphics[width=0.95\linewidth]{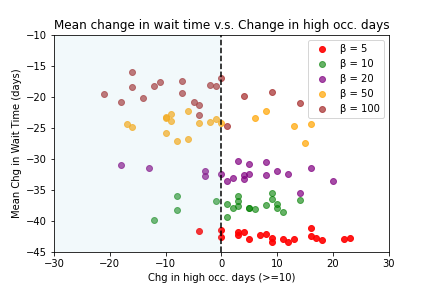}
         \caption{Mean change in wait times vs. ICU congestion}
         \label{fig:fig_sro_a-bi}
        \end{subfigure}%
        \begin{subfigure}{.5\textwidth}
        \centering
         \includegraphics[width=0.95\linewidth]{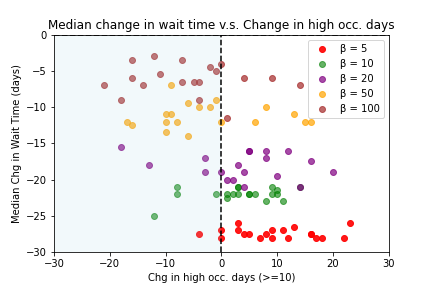}
         \caption{Median change in wait times vs. ICU congestion}
         \label{fig:fig_sro_b-bi}
    \end{subfigure}
    \caption{(Biweekly Scheduling) Performance trade-off of Standard-SRO between patient wait times and ICU congestion using different values of $\beta$ compared to the status quo. We also include greater values of $\beta$ to show that increasing $\beta$ further leads to longer wait times but insignificant reduction in ICU congestion.}
    \label{figs_sro-bi}
    \end{figure*}

       \begin{figure*}[!h]
        \centering
        \begin{subfigure}{.5\textwidth}
        \centering
         \includegraphics[width=0.95\linewidth]{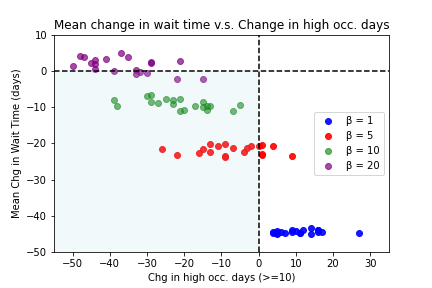}
         \caption{Mean change in wait times vs. ICU congestion}
         \label{fig:fig_csro_a-bi}
        \end{subfigure}%
        \begin{subfigure}{.5\textwidth}
        \centering
         \includegraphics[width=0.95\linewidth]{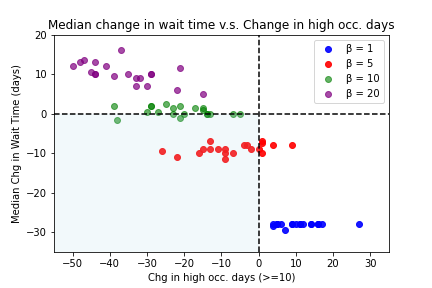}
         \caption{Median change in wait times vs. ICU congestion}
         \label{fig:fig_csro_b-bi}
    \end{subfigure}
    \caption{(Biweekly Scheduling) Performance trade-off of Conservative-SRO between patient wait times and ICU congestion using different values of $\beta$ compared to the status quo.}
    \label{figs_csro-bi}
    \end{figure*}

       \begin{figure*}[!h]
        \centering
        \begin{subfigure}{.5\textwidth}
        \centering
         \includegraphics[width=0.95\linewidth]{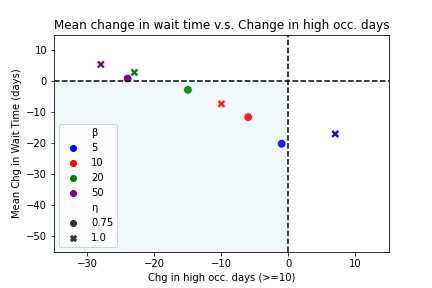}
         \caption{Mean change in wait times vs. ICU congestion}
         \label{fig:fig_robust_a-bi}
        \end{subfigure}%
        \begin{subfigure}{.5\textwidth}
        \centering
         \includegraphics[width=0.95\linewidth]{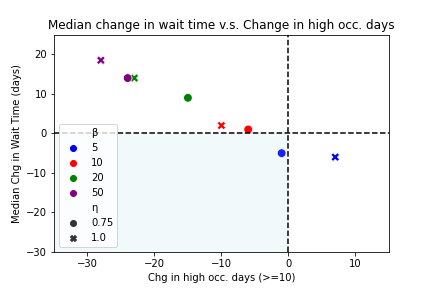}
         \caption{Median change in wait times vs. ICU congestion}
         \label{fig:fig_robust_b-bi}
    \end{subfigure}
    \caption{(Biweekly Scheduling) Performance trade-off of RRO between patient wait times and ICU congestion using different values of $\beta,\eta$ compared to the status quo. }
    \label{figs_robust-bi}
    \end{figure*}

\end{document}